\documentclass{article}

 \usepackage[preprint]{neurips_2026}

\usepackage[utf8]{inputenc} 
\usepackage[T1]{fontenc}    
\usepackage{hyperref}       
\usepackage{url}            
\usepackage{booktabs}       
\usepackage{amsfonts}       
\usepackage{nicefrac}       
\usepackage{microtype}      
\usepackage{xcolor}         

\usepackage{amsmath}
\usepackage{amssymb}
\usepackage{mathtools}
\usepackage{amsthm}
\usepackage{algorithm}
\usepackage{algorithmic}
\DeclareMathOperator*{\argmin}{arg\,min}
\DeclareMathOperator*{\argmax}{arg\,max}
\newcommand{\main}[1]{\textcolor{blue}{#1}}

\newtheorem{lemma}{Lemma}

\newtheorem{proposition}{Proposition}

\theoremstyle{plain}
\newtheoremstyle{TheoremNum}
    {\topsep}{\topsep}              
    {\itshape}                      
    {}                              
    {\bfseries}                     
    {.}                             
    { }                             
    {\thmname{#1}\thmnote{ \bfseries #3}}
\theoremstyle{TheoremNum}
\newtheorem{proposition-repeat}{Proposition}

\title{Batch Bayesian Active Learning \\ with Partial Batch Label Sampling}

\author{%
  Kangping Hu \\
  Department of Computer Science\\
  Georgia Institute of Technology\\
  Atlanta, GA 30332 \\
  \texttt{kangping.hu@gatech.edu} \\
  \And
  Stephen Mussmann \\
  Department of Computer Science\\
  Georgia Institute of Technology\\
  Atlanta, GA 30332 \\
  \texttt{mussmann@gatech.edu} \\
}

\begin{document}

\maketitle

\begin{abstract}
  Over the past couple of decades, many active learning acquisition functions have been proposed, leaving practitioners with an unclear choice of which to use. Bayesian active learning offers principled objectives with explainable intuition, including Expected Error Reduction (EER), Expected Predictive Information Gain (EPIG), and Bayesian Active Learning by Disagreements (BALD). A key challenge of such methods is the difficulty of extending to the batch setting, leading to either computational challenges (BatchBALD) or dramatic performance drops (top-$B$ selection). In this work, we extend active learning's Bayesian Decision Theory framework to derive Partial Batch Label Sampling (ParBaLS) for the EPIG algorithm. We show experimentally that ParBaLS EPIG gives superior performance for a variety of datasets with different budgets using Bayesian logistic regression on embeddings from large pre-trained models. Our code is available at \url{https://github.com/ADDAPT-ML/ParBaLS}.
\end{abstract}

\section{Introduction}

Active Learning (AL) aims to select the most informative data to label, given abundant unlabeled data but a limited labeling budget. The AL literature includes learning theory algorithms and analyses \citep{dasgupta2005coarse,balcan2006agnostic,balcan2007margin,zhang2014beyond,hanneke2015minimax,NEURIPS2025_e58dfd01}, heuristic methods \citep{lewis1995sequential, scheffer2001active, wang2014new, sener2018active, ash2019deep, pmlr-v162-zhang22k} and Bayesian methods \citep{houlsby2011bayesian, gal2017deep, kirsch2019batchbald, kirsch2021test, mussmann2022active, smith2023prediction}. While the learning theoretic branch is a rich literature with a variety of analytical insights, the algorithms are generally not used in practice. From a practical point of view, most heuristic active learning algorithms have inconsistent performance and lack explainability of when and why they work or not, making it hard for practitioners to decide which algorithm to use. They also require dataset-specific hyperparameters to balance tensions between principles such as uncertainty, diversity, and representativeness. Unlike model training, hyperparameter tuning is infeasible in active learning \citep{lowell2019practical,rauch2023activeglae,luth2023navigating}  because budget constraints prohibit collecting data with multiple hyperparameters.

\begin{figure*}[h]
\centerline{\includegraphics[scale=0.4]{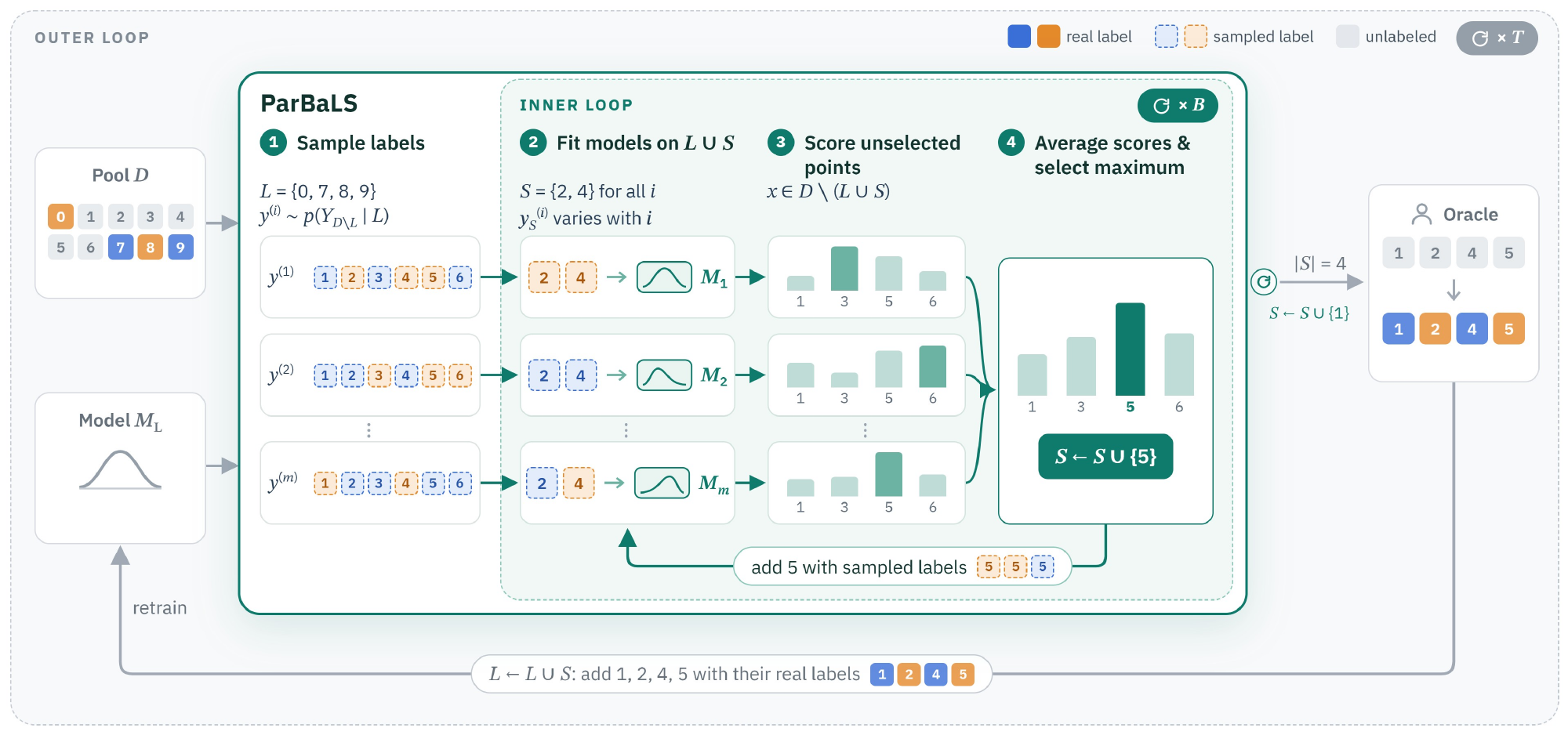}}
\caption{ParBaLS overview, with an example snapshot when the batch size is $B=4$, the currently labeled points are $L=\{7,8,9\}$, and the chosen but unlabeled points in the partial batch are $S=\{2,4\}$. We see how the point $\{5\}$ is selected by the inner loop and added to $S$. After this update, we assume another inner loop to select the point $\{1\}$ and achieve $|S|=4=B$. Then we execute the outer loop of sending the batch $S$ to the oracle annotator for ground-truth labels, which are added to $L$ for the next iteration.}
\label{fig:main}
\end{figure*}

On the other hand, Bayesian active learning algorithms use interpretable and direct principles, such as reducing test loss or parameter entropy. However, most objectives are naturally defined for single points, which often encounter batching as an obstacle, the key challenge that is addressed in this work. In modern machine learning, for practical reasons, data is typically labeled in batches rather than querying labels one-at-a-time. A common approach is to select the batch as the $B$ samples with the highest scores (top-$B$ selection), but this often leads to redundancy in selection, where similar high-scoring samples are selected together. Although \cite{kirsch2019batchbald} addresses the batching issue by extending the single-point objective of BALD to multiple points in a batch as BatchBALD, estimation of large joint probabilities can require exponentially many samples, making it infeasible even for moderate batch sizes. \cite{kirsch2023stochastic} lowers the computational cost for batching by adding Gumbel noise to the scores for selection, but our empirical results show that they are not as effective without hyperparameter tuning, which is not practical in active learning. With these limitations, it has often been found \citep{zhang2024labelbench, werner2024cross} that simpler heuristic algorithms like uncertainty sampling \citep{lewis1995sequential} work equally or better than principled Bayesian active learning algorithms.

We aim to avoid the redundancy of top-$B$ selection, the intractability of batch score estimation, and the need to annotate one data point at a time. Batches are required for practicality, but single-point selection is typically more effective. Our key insight is to keep the outer loop as batch active learning, while simulating labeling one data point every step in the inner loop using pseudo-labels. In this case, each data point selection aligns with the single-point objective, while the annotation process aligns with batch active learning.

Here, Partial Batch Label Sampling (ParBaLS) is derived from the Bayesian Decision Theory (BDT) principle: choose the action (point to label) that minimizes the expected cost (the test loss). In particular, our method incrementally builds a partial batch one-at-a-time. Since we don't have ground-truth labels for the partial batch, we use Monte Carlo sampled pseudo-labels to update the model within this inner loop. We outline ParBaLS in Figure \ref{fig:main}. We show experimentally that ParBaLS with the EPIG criterion (equivalent to reducing expected test loss for the negative log likelihood loss \citep{mussmann2022active}) has the uniformly best performance against other Bayesian batching baselines and the best performance among a set of heuristic baselines. In particular, we compare active learning batching approaches with Bayesian logistic regression on pre-trained embeddings across 11 tabular, text, and image datasets under three budget settings, covering native multiclass classification \citep{Kelly2025UCI}, binary label imbalance, and natural distribution shifts \citep{koh2021wilds}.

Our main contributions are twofold:

\begin{itemize}
    \item We introduce Partial Batch Label Sampling (ParBaLS) for the challenge of batching in Bayesian active learning.
    \item We demonstrate the effectiveness of ParBaLS by conducting experiments with Bayesian Logistic Regression on tabular, text, and image datasets with embeddings from large pre-trained models.
\end{itemize}

\section{Background \& Setting}\label{background}

\subsection{Batch Active Learning for Classification}\label{sec:active}

We assume an input domain $\mathcal{X}$, output domain $\mathcal{Y}$, and a data distribution $\mathcal{D}$ over $\mathcal{X} \times \mathcal{Y}$. We decompose $\mathcal{D}$ into the unlabeled distribution $\mathcal{D}_X$ and the conditional label distribution $\mathcal{D}_{Y\mid X}$. We assume we have an unlabeled pool dataset sampled from $\mathcal{D}_X$, which we call $D$. We assume the pool-based active learning setting, where we can select points from $D$ and observe their label sampled according to $\mathcal{D}_{Y \mid X}$. The goal of Batch Active Learning is, given $D$ and an initial labeled seed set $L_0$, iteratively select $T$ batches, $S_1,S_2,\dots,S_T\subset D$ of size $|S_t|=B$ for labeling, so that after training on the labeled set, the test loss is low.

We assume a training procedure which, given a labeled dataset $L$, computes a model $M_L$ that predicts a probability distribution over the labels of the (unlabeled) datapoints. We outline a batching algorithmic template as Algorithm \ref{alg:al} in Appendix \ref{app:explanation}. Following \citet{mussmann2022active} and \citet{smith2023prediction}, we assume access to an \emph{unlabeled} validation set $V \subset \mathcal{X}$ that is drawn from the target distribution, which can also be drawn from the unlabeled pool dataset if there is no distribution shift.

\subsection{Bayesian Decision Theory and Unbatched Active Learning}
\label{sec:bdt}

For now, assume that we are in the unbatched setting with $B=1$. We assume a Bayesian model with a prior joint distribution over all labels $Y_x$, where $Y_x$ is a random variable for the label corresponding to a datapoint $x$. For notational convenience, we also use the random variable to denote its distribution as in: $\mathbb{E}_{y_x \sim Y_x}[y_x]$. After conditioning on training labels, we have a posterior joint distribution over the unlabeled points.

Practically, a Bayesian distribution over labels is typically implemented as a Bayesian parametric model, where labels are conditionally independent given the model parameters. In this case, the posterior over labels can be efficiently computed by Bayesian model averaging over samples from the parameter posterior, where the model $M_L$ is ``trained'' by sampling and saving parameters from the posterior.

Bayesian Decision Theory \citep{parmigiani2009decision} provides a formal framework for making optimal decisions under uncertainty. It is grounded in the principles of probability theory and utility theory, combining prior beliefs with observed data to guide rational decision-making. At its core, the theory assumes that all uncertainty can be quantified probabilistically, and that decisions should be made to minimize expected cost. Bayesian Decision Theory chooses the action that minimizes the expected cost, where the expectation is taken according to the Bayesian model.

In our setting, we define the final action as predicting a probability vector $p \in \Delta_\mathcal{Y}$ for points drawn from a target distribution, where $\Delta_\mathcal{Y}$ is the probability simplex over the labels $\mathcal{Y}$. Our goal is to incur a low expected loss $\ell(y,p)$. We use the negative log likelihood as the loss, $\ell(y,p) = - \ln p[y]$, where $p[y]$ is the entry of $p$ corresponding to $y\in\mathcal{Y}$.

For active learning, before the final action of predicting on target points, we have $T$ actions corresponding to selecting batches of unlabeled data points $S_1, \dots, S_T$ for labeling. Due to the complexity of planning over a long sequence of actions, following the vast majority of Bayesian active learning methods \citep{houlsby2011bayesian, gal2017deep, kirsch2021test, mussmann2022active, smith2023prediction}, we focus on myopic decisions. In particular, we choose each batch as if it were the last batch before the prediction action.

Formally, given a labeled set $L \subset \mathcal{X}$ (with corresponding labels $y_L \in \mathcal{Y}^{|L|}$), our first action is to choose an additional data point $\hat{x}$ from $D$, we then observe label $\hat{y}$, then our second action is to make probabilistic predictions $P \in \Delta_\mathcal{Y}^{|V|}$ on validation points $V \subset \mathcal{X}$. With $y_V \in \mathcal{Y}^{|V|}$, the expected cost if we label $\hat{x}$ is
\begin{equation}
\begin{aligned}
\label{eq:bdt-nonbatch}
    \mathrm{Cost}_{L,V,y_L}(\hat{x})= \mathbb{E}_{\hat{y} \sim Y_{\hat{x}}|Y_L=y_L}\left[ \min_{P \in \Delta_\mathcal{Y}^{|V|}} \mathbb{E}_{y_V \sim Y_V | Y_{\hat{x}}=\hat{y},Y_L=y_L}\left[\frac{1}{|V|}\sum_{x \in V} \ell(y_x,P_x)\right]\right]
\end{aligned}
\end{equation}

and so the next point to label according to Bayesian Decision Theory is $\argmin_{\hat{x} \in D} \mathrm{Cost}_{L,V,y_L}(\hat{x})$.

For the negative log loss $\ell(y,p) = - \ln p[y]$, the optimal $P$ is simply the posterior distribution \citep{mussmann2022active}, and it can be shown that,

\begin{proposition}
\label{prop:epig-derive}
If $\ell(y,p) = - \ln p[y]$, then there is a constant $c$ that doesn't depend on $\hat{x}$ such that 
\begin{align}
    \mathrm{Cost}_{L,V,y_L}(\hat{x}) =  c - \frac{1}{|V|} \sum_{x \in V} I(Y_x; Y_{\hat{x}}|Y_L = y_L)
\end{align}
\end{proposition}

For completeness, the proof is in Appendix \ref{app:derive-proof}. This implies that minimizing the expected cost is equivalent to maximizing the average mutual information with the validation points. This is the same form as Expected Predictive Information Gain (EPIG) \citep{kirsch2021test, smith2023prediction} (motivated by predictive mutual information), EER (motivated by reducing expected error, see derivation in \cite{mussmann2022active}), and a method from one of the first active learning papers \citep{10.1162/neco.1992.4.4.590}.
By focusing directly on minimizing predictive error, they are more robust and interpretable than heuristic algorithms.

\section{Method}\label{sec:parbals}

In Section \ref{sec:bdt}, we reviewed how the the EPIG criterion was derived from Bayesian Decision Theory where the cost is the negative log likelihood on an unlabeled validation set. We now turn towards the issue of extending this method to batch active learning.

We accomplish this by decomposing the choice of a batch into $B$ actions of selecting individual points. Then, similar to BDT for unbatched active learning, we can efficiently compute the myopically optimal action, conditioned on the points we've already committed to selecting for the batch, but haven't observed the labels for. We refer to the points $S$ that we've selected but haven't labeled as the \emph{partial batch}.

Let $Y_S$ be a random variable (with values in $\mathcal{Y}^{|S|}$) for the labels corresponding to the unlabeled points $S$. For $L$ labeled points and a partial batch of $S$ points, the expected cost if we select point $\hat{x}$ is
\begin{equation}
\begin{aligned}
\label{eq:parbals-cost}
   \mathrm{Cost}_{L,S,V,y_L}(\hat{x}) =\mathbb{E}_{\hat{y},y_S \sim Y_{\hat{x}},Y_S|Y_L=y_L} \left[ \min_{P \in \Delta_\mathcal{Y}^{|V|}} \mathbb{E}_{y_V \sim Y_V | Y_{\hat{x}}=\hat{y},Y_S=y_S,Y_L=y_L}\left[\frac{1}{|V|}  \sum_{x \in V} \ell(y_x,P_x)\right]\right]
\end{aligned}
\end{equation}

The (myopically) optimal point according to Bayesian Decision Theory is $\argmin_{\hat{x} \in D} \mathrm{Cost}_{L,S,V,y_L}(\hat{x})$. It is straightforward to show (proof in Appendix~\ref{app:derive-proof}) that,
\begin{proposition}
\label{prop:parbals}
If $\ell(y,p) = - \ln p[y]$, then there is a constant $c$ that doesn't depend on $\hat{x}$ such that
\begin{equation}
\begin{aligned}
    \mathrm{Cost}_{L,S,V,y_L}(\hat{x}) =  c - \mathbb{E}_{y_S \sim Y_S|Y_L=y_L} \left[\frac{1}{|V|}  \sum_{x \in V} I(Y_x; Y_{\hat{x}}| Y_S = y_S,Y_L=y_L) \right]
\end{aligned}
\end{equation}
\end{proposition}

Thus, minimizing the cost is equivalent to maximizing the expected mutual information. A computation challenge is the expectation over $Y_S$. To avoid summing over all $|\mathcal{Y}|^{|S|}$ values of $y_S$, we use a Monte Carlo estimate by sampling $m$ versions of $y^{(i)} \sim Y_D | Y_L = y_L$, where the labels are \emph{dependent} within each of the $m$ samples but \emph{independent} across the $m$ samples. We call each sample a \emph{universe}. We select the next point to add to the partial batch as,
\begin{equation}
    \begin{aligned}\label{eq:obj}
        \mathrm{ParBaLS}(D,L,S,V,y_L,\{y^{(i)}\}_{i=1}^m) \in \argmax_{\hat{x} \in D} \sum_{i=1}^m \sum_{x \in V} I(Y_x; Y_{\hat{x}}| Y_S = y^{(i)}_S,Y_L=y_L)
    \end{aligned}
\end{equation}

We refer to $y^{(i)}$ as \emph{pseudo-labels} since they are sampled from the Bayesian model rather than labeled by the oracle annotator. 
Since we use sampled labels for the partial batch, we refer to our method as Partial Batch Label Sampling (ParBaLS). We can easily show that the objective maximized by ParBaLS has expectation $\mathrm{Cost}_{L,S,V,y_L}$ and that the expected sub-optimality of ParBaLS's selected point decays as $\mathcal{O}\left(\frac{1}{\sqrt{m}}\right)$. See Appendix~\ref{app:proof-mc-approx} for the precise statement and the proof.

A key advantage is that we can incrementally train $m$ parallel models $\{M_i\}_{i=1}^m$ on $y^{(i)}_S$ to avoid the issue of estimating joint probabilities over $B+1$ labels (exponential in the batch size). Precisely, we assume we can train a model $M$ (e.g., sample sets of Bayesian posterior parameter samples) on labeled samples $(L,y_L)$, in order to have efficient access to (approximate) probabilities $\Pr_M(Y_S = y_S) = \Pr(Y_S=y_S|Y_L = y_L)$ and samples $Y_D \sim M$ (approximately) equivalent to sampling from $Y_D | Y_L = y_L$. With $M_i$ as a model trained on $(L \cup S, y_L \cup y_S^{(i)})$, we can equivalently define ParBaLS with $\mathrm{ParBaLS}(D ,V,\{M_i\}_{i=1}^m)$ as
\begin{equation}
    \begin{aligned}
    \label{eq:alg}
        \argmax_{\hat{x} \in D} \sum_{i=1}^m \sum_{x \in V} \sum_{\hat{y} \in \mathcal{Y}} \sum_{y \in \mathcal{Y}} \Pr_{M_i}(Y_{\hat{x}}=\hat{y},Y_x=y) \cdot\log\left( \frac{\Pr_{M_i}(Y_{\hat{x}}=\hat{y},Y_x=y)}{\Pr_{M_i}(Y_{\hat{x}}=\hat{y}) \Pr_{M_i}(Y_x=y)} \right)
    \end{aligned}
\end{equation}

We outline ParBaLS EPIG in Algorithm \ref{alg:parbals}. We additionally include a diagram comparing the standard AL workflow (Figure~\ref{fig:active}) and the ParBaLS workflow (Figure~\ref{fig:parbals}), and another illustrative diagram (Figure \ref{fig:example}) in Appendix \ref{app:explanation}. 

\subsection{Conceptual Comparison}
\label{sec:conceptual-comparison}

Conceptually, the myopic construction of a batch is similar to BatchBALD. Although BatchBALD is for BALD and focuses on parameter uncertainty, a similar version could be derived for EPIG to focus on predictive uncertainty. However, two key differences are that BatchBALD uses exhaustive enumeration of the joint label set (though the paper mentions the possibility of Monte Carlo importance sampling) and the intractability of estimating large joint probabilities. Intuitively, the number of posterior parameter samples needed to accurately estimate the joint probability distribution over $B$ labels scales exponentially in $B$. ParBaLS avoids this, only requiring the computation of joint probabilities over two random variables. More details are provided in Appendix~\ref{sec:batch}.

Two other less related ideas are BAIT \citep{ash2021gone} and ACS-FW \citep{pinsler2019bayesian}. However, while ParBaLS EPIG is derived by Bayesian Decision Theory, ACS-FW is derived by preserving the \emph{expected} log posterior, which has an unclear connection to predictive performance. BAIT is derived by optimizing the predictive performance in a frequentist framework, however it relies on asymptotic statistics which may not hold for the small data budgets that typically motivate active learning.

\subsection{Computing ParBaLS}
\label{sec:computation}
\paragraph{Computational Complexity} Treating the model retraining time as $R$, the computational complexity to select a batch is $\mathcal{O}\left( B \cdot m \cdot (R + |D| \cdot |V| \cdot |\mathcal{Y}|^2)\right)$. For comparison, naive top-$B$ batching for EPIG (and retraining after the batch) has time complexity $\mathcal{O}\left(R + |D| \cdot |V| \cdot |\mathcal{Y}|^2\right)$. Thus, ParBaLS adds a factor of $B \cdot m$. While this factor can be reasonably large (e.g., a hundred), the re-training time $R$ could be significantly diminished by leveraging incremental re-training rather than repeatedly training from scratch, since only a single point is added to $M_i$ for re-training. Notably, the scaling is quadratic in the number of classes, making EPIG (both top-$B$ and ParBaLS) slow for many classes. To address this, we mostly focus on binary datasets experimentally.

\paragraph{Scalability} The $m$ universes are conditionally independent and can be updated in parallel. ParBaLS is particularly practical when labels are acquired in large batches and acquisition uses frozen embeddings with a Bayesian last layer: the expensive encoder is evaluated once, while each hypothetical update adds only one point to a linear classifier. Thus, the method is scalable to settings where incremental and parallel computation are available.

\paragraph{ParBaLS-MAP} While the main version of ParBaLS requires sampling $m$ versions of $y^{(i)}$, another variant we test experimentally involves using just one version of labels $y^{(1)}$ comprised of the Maximum A Posteriori (MAP) labels for each unlabeled point: $y^{(1)}_x = \argmax_{y \in \mathcal{Y}} \Pr(Y_x = y | Y_L=y_L)$. This decreases the computational complexity by a factor of $m$. This is also more similar to the usual definition of pseudo-labels as the current model's most likely prediction. However, the objective optimized by this variant is no longer an unbiased estimate of the true cost, and we see in the experiments that this variant is outperformed by ParBaLS.

\begin{algorithm}[tb]
\caption{ParBaLS EPIG}
\label{alg:parbals}
\renewcommand{\algorithmicrequire}{\textbf{Input:}}
\renewcommand{\algorithmicensure}{\textbf{Output:}}
\begin{algorithmic}[1]
\REQUIRE unlabeled pool $D$; unlabeled validation set $V$; currently labeled subset $L$ with labels $y_L$; Bayesian model training procedure $\mathrm{Train}$; budget for each iteration $B$; number of universes $m$
\ENSURE Batch $S$ for labeling
\STATE Train Bayesian model $M_L=\mathrm{Train}(L,y_L)$
\FOR{$i \leftarrow 1$ {\bfseries to} $m$}
\STATE Sample an independent set of intra-dependent pseudo-labels $y^{(i)} \sim Y_D | Y_L = y_L$ using $M_L$
\STATE Initialize $M_i = M_L$
\ENDFOR
\STATE $S \leftarrow \emptyset$
\FOR{$j \leftarrow 1$ {\bfseries to} $B$}
    \STATE Compute $\hat{x} = \mathrm{ParBaLS}(D,V,\{M_i\}_{i=1}^m)$ according to Equation~\ref{eq:alg}
    \STATE $S \leftarrow S \cup \{\hat{x}\}$
    \STATE $D \leftarrow D \setminus \{\hat{x}\}$
    \STATE For each $i \in [m]$, update $M_i = \mathrm{Train}(L \cup S, y_L \cup y^{(i)}_S)$
\ENDFOR
\STATE \textbf{return} $S$
\end{algorithmic}
\end{algorithm}

\section{Experiments}\label{exp}

\subsection{Datasets}

We evaluate on three image classification datasets: CIFAR-10 \citep{krizhevsky2009learning}, iWildCam \citep{beery2021iwildcam}, and fMoW \citep{christie2018functional}; three text classification datasets: AG News \citep{agnews2004}, Yelp Review Full \citep{zhang2015character}, and Civil Comments \citep{DBLP:journals/corr/abs-1903-04561}; two binary tabular datasets from Kaggle: Airline Passenger Satisfaction \citep{airline2020} and Credit Card Fraud \citep{credit2022}; and three tabular datasets from UCI \citep{Kelly2025UCI}: Magic Gamma Telescope (binary), Statlog Landsat Satellite (6 classes), and Vowels (11 classes). iWildCam and fMoW are from the WILDS benchmark \citep{koh2021wilds} and contain natural distribution shifts.

As mentioned in Section~\ref{sec:computation}, EPIG (and its batching variants) does not scale well to a large number of classes. To address this, we convert the image and text datasets to one-vs-all. This has the side effect of creating the  common practical challenge of label imbalance where data selection matters more \citep{ertekin2007learning,attenberg2013class,mussmann2020importance}. Specifically, the first original class is the positive class, and all other original classes are the negative class. We leave the tabular dataset classes untouched, including two multiclass datasets with 6 and 11 classes. 

\subsection{Models}

\subsubsection{Bayesian Logistic Regression}

Past work has shown that linear models on fixed pre-trained embeddings are competitive for active learning using selection-via-proxy \citep{zhang2024labelbench}. Given that recent work has found Bayesian neural networks exhibit non-Bayesian properties \citep{pmlr-v267-pituk25a}, we focus on Bayesian linear models on pre-trained embeddings. This benefits from the growing availability of effective embedding models, achieving both efficiency and performance. We provide the detailed setup for Bayesian logistic regression (BLR) in Appendix \ref{sec:setup}. 

\subsubsection{Encoders}
For embeddings, we use CLIP-ViT-B/32 \citep{radford2021learning} for the WILDS datasets, DINOv2-ViT-S/14 model \citep{oquab2024dinov2} for CIFAR-10, and BERT \citep{devlin2019bert} for the text datasets, which are widely used in recent studies \citep{zhang2024labelbench, huseljic2024fast}. The choices of the vision encoders (CLIP vs DINO) are based on the better-performing encoder in preliminary random selection experiments. We provide additional details about feature preprocessing in Appendix \ref{sec:setup}. 

\subsection{Active Learning Algorithms} 

For Bayesian active learning baselines, we evaluate EPIG \citep{smith2023prediction} and BALD \citep{houlsby2011bayesian, gal2017deep}, where each of them is combined with top-$B$ batch selection, one of the three Gumbel noise methods \citep{kirsch2023stochastic}, or BatchBALD \citep{kirsch2019batchbald} for BALD. For non-Bayesian heuristic methods, we include Random, Confidence \citep{lewis1995sequential}, GLISTER \citep{killamsetty2021glister}, CoreSet \citep{sener2018active}, BADGE \citep{ash2019deep}, and GALAXY \citep{pmlr-v162-zhang22k} as our baselines, which are selected from top-performing algorithms in \cite{zhang2024labelbench} and other state-of-the-art active learning papers. For fair comparison, we implement the heuristics with the calibrated, more accurate probabilities from our Bayesian logistic regression model. Our proposed algorithms are EPIG with our ParBaLS-MAP and ParBaLS. We use $m=10$ label samples for ParBaLS in our main experimental results, and conduct ablations over $m$ in Appendix \ref{ablations}. We discuss more details about experimental setup, metrics, and statistical interpretation in Appendix \ref{sec:setup}.

\subsection{Experimental Results}

\begin{table*}
\centering
\resizebox{\columnwidth}{!}{%
\begin{tabular}{l|cccc|cccc}
\hline
Algorithm & \multicolumn{4}{c|}{Bayesian Tables} & \multicolumn{4}{c}{Heuristics Tables} \\
& \underline{\textbf{Highest Mean (HM)}} & \textbf{Ties with HM} & Wins $-$ Losses & Highest Loss Rate$\downarrow$ & \underline{\textbf{Highest Mean (HM)}} & \textbf{Ties with HM} & Wins $-$ Losses & Highest Loss Rate$\downarrow$ \\ \hline
Random & - & - & - & - & 1.65\% & 36.09\% & -1.08 & 61.71\% \\
Confidence & - & - & - & - & 14.33\% & 71.35\% & +1.56 & 25.62\% \\
GLISTER & - & - & - & - & 2.20\% & 26.72\% & -3.07 & 75.76\% \\
CoreSet & - & - & - & - & 1.93\% & 28.93\% & -4.03 & 71.90\% \\
BADGE & - & - & - & - & 2.75\% & 39.12\% & -0.95 & 61.98\% \\
GALAXY & - & - & - & - & 26.45\% & 72.73\% & +1.92 & 23.97\% \\
\hline
Top-$B$ BALD & 1.10\% & 33.33\% & -5.34 & 74.38\% & - & - & - & - \\
PowerBALD & 0.83\% & 50.96\% & -0.14 & 51.79\% & - & - & - & - \\
SoftmaxBALD & 4.13\% & 41.87\% & -1.29 & 62.26\% & - & - & - & - \\
SoftRankBALD & 3.03\% & 40.77\% & -2.01 & 64.19\% & - & - & - & - \\
BatchBALD & 1.65\% & 51.24\% & -2.71 & 65.29\% & - & - & - & - \\
\hline
Top-$B$ EPIG & 4.41\% & 68.32\% & +1.77 & 47.11\% & - & - & - & - \\
PowerEPIG & 3.31\% & 42.98\% & -1.63 & 63.64\% & - & - & - & - \\
SoftmaxEPIG & 2.20\% & 46.56\% & -0.50 & 55.92\% & - & - & - & - \\
SoftRankEPIG & 7.44\% & 64.46\% & +2.01 & 41.32\% & - & - & - & - \\
\hline
ParBaLS-MAP EPIG & 14.60\% & 93.39\% & +4.54 & 9.64\% & 11.57\% & 77.41\% & +2.61 & 16.53\% \\
\textbf{ParBaLS EPIG} & \textbf{58.40\%} & \textbf{99.72\%} & \textbf{+5.29} & \textbf{1.38\%} & \textbf{39.12\%} & \textbf{87.60\%} & \textbf{+3.04} & \textbf{14.88\%} \\
\hline
\end{tabular}
}
\caption{Leaderboard across all 11 datasets and three budget settings, evaluated separately at each iteration. Highest Mean (HM) and Ties with HM show the percentages of available dataset-budget-iteration evaluations. Wins $-$ Losses is the sum of normalized pairwise win rates in a strategy's Figure \ref{fig:win-matrices} row minus the corresponding column sum. Highest Loss Rate is the worst loss rate of each algorithm in Figure \ref{fig:win-matrices}, where lower is better. BatchBALD is evaluated only for the $B=10$ budget setting.}
\label{tab:alg_ttest}
\end{table*}

\begin{figure*}[h]
\centerline{\includegraphics[scale=0.35]{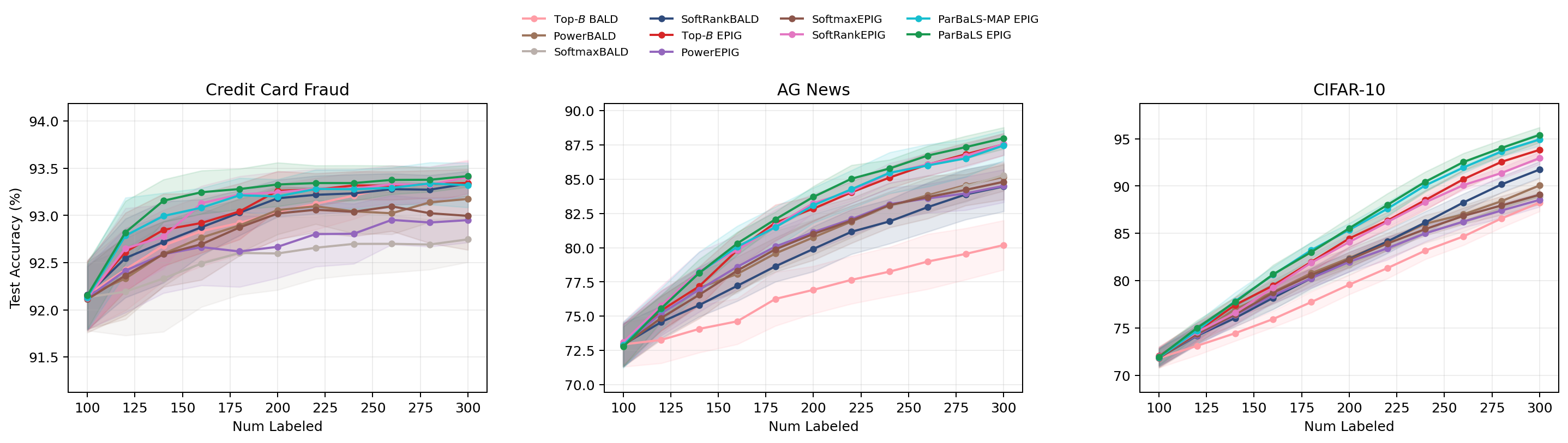}}
\caption{Learning curves of Bayesian AL algorithms with Bayesian Logistic Regression, where each of the 10 iterations has a labeling budget of 20 samples, starting with random initialization of 100 samples.}
\label{fig:bs100_bayesian}
\end{figure*}

\begin{figure*}[h]
\centerline{\includegraphics[scale=0.35]{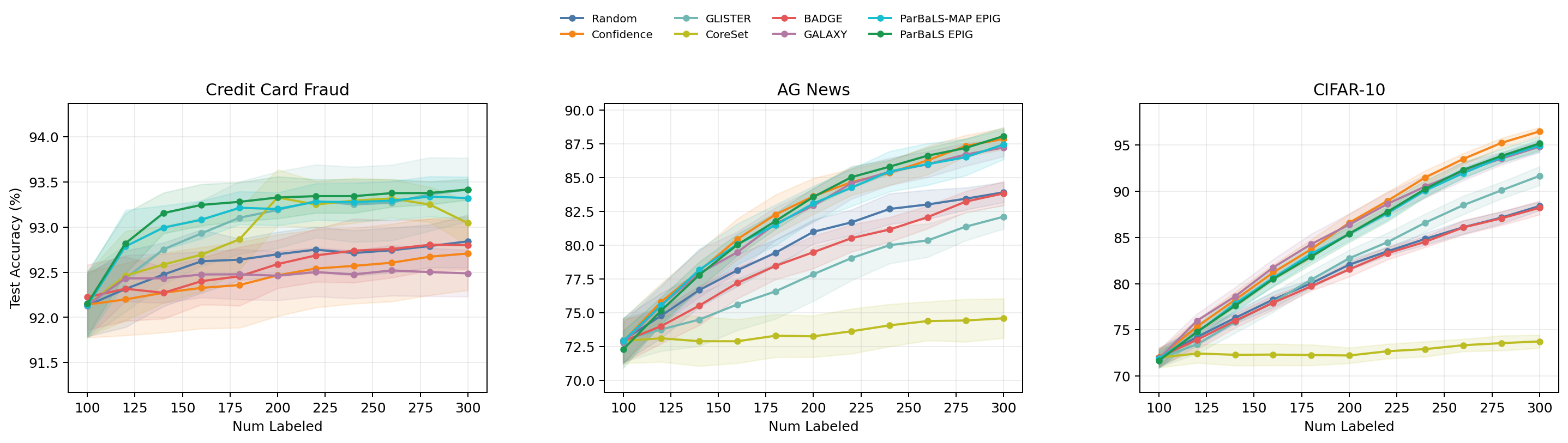}}
\caption{Learning curves of ParBaLS and heuristics baselines with Bayesian Logistic Regression, where each of the 10 iterations has a labeling budget of 20 samples, starting with random initialization of 100 samples.}
\label{fig:bs100_heuristic}
\end{figure*}

We run experiments with three data budget settings, each with $T=10$ iterations. The first uses an initial labeled set of $100$ and a batch size of $20$. A comparison among Bayesian approaches is shown in Figure \ref{fig:bs100_bayesian} and a comparison of our methods to heuristic approaches is shown in Figure \ref{fig:bs100_heuristic} on a selected tabular, text, and image dataset. We use two additional budget settings, one with only $20$ initial datapoints and one with a batch size of $10$. Learning curves for the three selected datasets are shown in Appendix~\ref{sec:curves}. The tables with the results for all datasets are shown in Appendix~\ref{sec:tables}.

Due to its poor scaling in the batch size, BatchBALD timed out without completing one iteration given the time required by ParBaLS to complete all iterations for $B=20$. Thus, numbers are only presented for BatchBALD in the $B=10$ budget setting, where it is not a top performer. 

Using 11 datasets, 3 budgets, and 10 iterations, there are 330 different settings. First, for each algorithm, we can compute the proportion of settings where the algorithm had the Highest Mean (HM). Second, for each algorithm, we compute a statistical-significance test across the 10 seeds against the method with HM, and count the proportion where the performance is not statistically significantly different from HM. Next, for each pair of methods, we compute Win-Tie-Loss numbers across the settings, where a Tie is counted when the performance of the two methods are not statistically significant. The full Win-Tie-Loss matrices are shown in Figure~\ref{fig:win-matrices} of Appendix~\ref{sec:tables}. For each method, we report the difference between the sum of all wins (against other methods) and the sum of all losses. We further report the highest loss rate for each method as a measure of robustness. The four summary metrics for both the Bayesian methods and the heuristic methods are shown in Table~\ref{tab:alg_ttest}.

In summary, ParBaLS EPIG is uniformly best compared to the Bayesian methods where it is tied for best 99.72\% of the time, and loses at most 1.38\% of the time against other approaches. ParBaLS-MAP EPIG is somewhat worse, but better than the Bayesian baselines. The next best Bayesian methods are Top-$B$ EPIG and SoftRankEPIG, but they have a loss rate of at least 40\%.

Compared to the heuristic methods, ParBaLS EPIG is the strongest algorithm, though it isn't uniformly the best. It is tied with the HM in 87.60\% of settings and loses around 15\% of the time to the best competitor. ParBaLS-MAP EPIG is a step down, but still better than the top baselines, Confidence and GALAXY, in most scenarios.

The use of the significance test is uncorrected for multiple hypotheses. Thus, we further provide a paired global statistical comparison between ParBaLS EPIG and Top-$B$ EPIG in Table \ref{tab:global-paired} in Appendix \ref{sec:tables}, that shows ParBaLS significantly improves over Top-$B$ EPIG for final test accuracy, iteration-averaged test accuracy, final negative log likelihood, and iteration-averaged negative log likelihood.

\begin{figure*}[h]
\centerline{\includegraphics[scale=0.35]{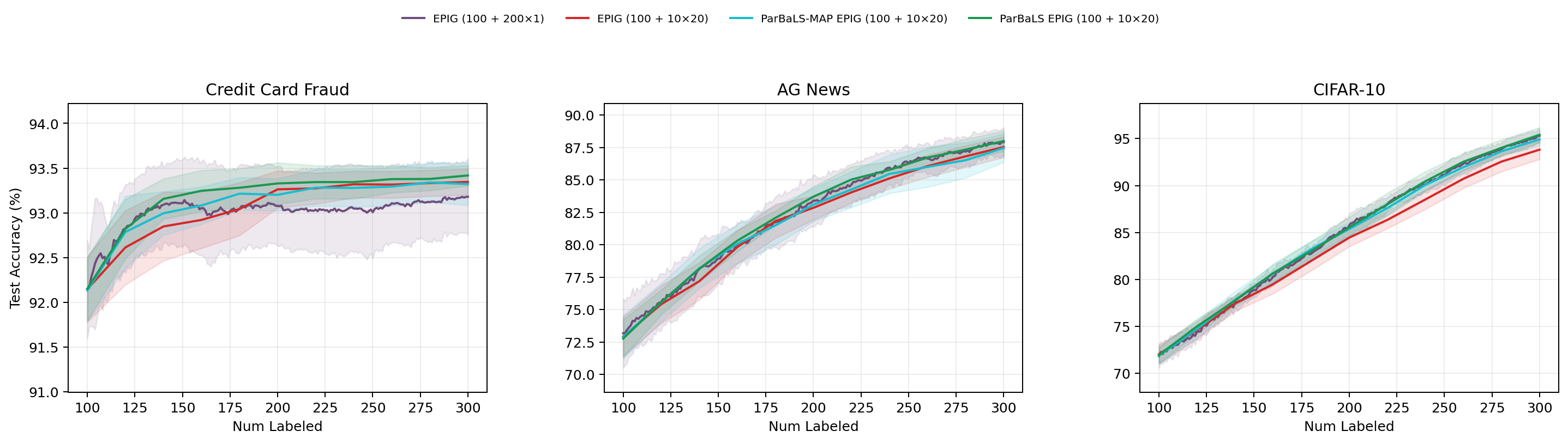}}
\caption{Comparison between batch acquisition and single selection with a total of 300 labeling budget, starting with random initialization of 100 samples.}
\label{fig:200by1}
\end{figure*}

Considering the EPIG objective is designed for single selection, we conduct $T = 200$ iterations with each iteration budget of $B = 1$, starting with random initialization of 100 samples, as shown in Figure \ref{fig:200by1}. We can see that ParBaLS (with $B=20$) mostly matches the $B=1$ algorithm while outperforming Top-$B$ selection (with $B=20$). This experiment is only for a conceptual comparison since single selection can be impractical in modern machine learning applications, as it requires a large number of data collection iterations. For example, with human annotation, hundreds of single selection iterations are difficult to crowd-source. Even with LLM-based annotation, batch querying is generally cheaper and more efficient.

Finally, we evaluate multiple choices of the positive class in Appendix~\ref{ova_ood}.

\section{Related Work}

\subsection{Bayesian Batching in Active Learning}

We categorize existing Bayesian AL batching strategies into three groups:

\paragraph{Top-$B$} Compute the score for each point in the pool, then take the $B$ points with the highest scores. Unfortunately, this method ignores the dependencies between batch labels and can be very sub-optimal. For example, this strategy may select a batch composed of similar data points that can be all informative at that given time, but redundant as a batch.
See \cite{kirsch2019batchbald} for more intuition and examples.

\paragraph{Heuristic Diversity} Heuristically add randomness or diversity to the batch. For example, \cite{kirsch2023stochastic} randomly perturbs the scores and takes the top-$B$, and \cite{Wei2015SubmodularityID} uses submodular maximization to pick a diverse subset of the top-scoring points. Such methods unfortunately require tuning dataset-specific hyperparameters, which are infeasible to tune for active learning, where we collect data once. Though \cite{kirsch2023stochastic} finds that a default hyperparameter ($\beta=1$) works well uniformly, we arrive at a different conclusion in our experiments. Intuitively, the level of randomness depends on the level of statistical dependency between high-scoring points, which is dataset-specific.

\paragraph{Greedy Batch Acquisition} We can replace optimizing $\hat{x} \in D$ with optimizing a batch of points $\hat{X} \subset D$ with $|\hat{X}|=B$. BatchBALD \citep{kirsch2019batchbald} optimizes $\hat{X}$ via greedy subset selection. As described in Section~\ref{sec:conceptual-comparison}, estimating joint probabilities over many variables is computationally intractable.

\subsection{Bayesian Optimization}

A line of work broadly related to active learning is Bayesian optimization, where the goal is to find an optimizer, rather than to maximize prediction accuracy. In parallel Bayesian optimization, batches are constructed before their outcomes are observed. Kriging Believer and Constant Liar \citep{ginsbourger2010kriging} build a batch greedily, imputing a single deterministic outcome for each pending point and updating the Gaussian-process surrogate as if it had been observed. Multi-step Bayesian optimization \citep{jiang2020efficient} draws multiple sampled hypothetical outcomes to estimate the value of future sequential decisions in a lookahead tree, rather than to account for pending evaluations within a batch. Deterministic imputation is most closely related to our ParBaLS-MAP ablation, whereas ParBaLS samples multiple joint label universes, scores points using a separately conditioned classifier posterior for each, and averages their prediction-oriented utilities. By sampling, ParBaLS yields unbiased estimates of the cost and a finite-$m$ decision guarantee.

\subsection{Pseudo-labeling in Active Learning}

Some algorithms combine pseudo-labeling, a common semi-supervised learning technique, with active learning to handle the issue of a limited labeling budget from complementary directions \citep{tharwat2023survey}. For example, \citet{zhu2003combining} performs active learning on top of the pseudo-labeled samples, reducing the impact of the uneven quality of the pseudo labels. On the other hand, \citet{gao2020consistency} and \citet{ji2025semi} combine several heuristic algorithms and use semi-supervision as the learner. \citet{kirsch2021test} and \citet{10.5555/3625834.3625999} train a model with both real-labeled train data and pseudo-labeled evaluation data, which is used to compute acquisition scores for unlabeled data. In contrast, ParBaLS introduces a fundamentally different use of pseudo-labeling, where the pseudo-labels have multiple versions and are temporary, replaced by ground-truth labels after a batch is constructed. This use of pseudo-labels is more similar to \citet{jiang2017efficient}, which applies sequential simulation to facilitate batch active search.

\section{Conclusion}

In this work, we derive ParBaLS, a batched active learning algorithm from the Bayesian Decision Theory principle. ParBaLS tackles the batch acquisition issue of existing Bayesian active learning algorithms by performing active learning with one-at-a-time queries but with sampled pseudo-labels. We show that ParBaLS EPIG overall outperforms other algorithms in 33 data settings. We foresee a broad opportunity for future work to discover computationally efficient approximations to scale the ParBaLS EPIG algorithm to larger datasets and models. In particular, incremental Bayesian learning methods could leverage the fact that the datasets grow by just one point at a time.

\bibliography{reference}
\bibliographystyle{plainnat}


\appendix

Here, we first provide more conceptual formulation and illustration in Appendix \ref{app:explanation}. Then, we elaborate on theoretical derivations and comparison in Appendix \ref{opt} and Appendix \ref{sec:batch}, respectively. After that, we present detailed experimental setup and results in Appendix \ref{sec:results}. Finally, we enumerate the license information of related assets in Appendix \ref{sec:licenses}.

\section{Active Learning Formulation and ParBaLS Illustration}
\label{app:explanation}

\begin{figure*}[h]
\centerline{\includegraphics[scale=0.4]{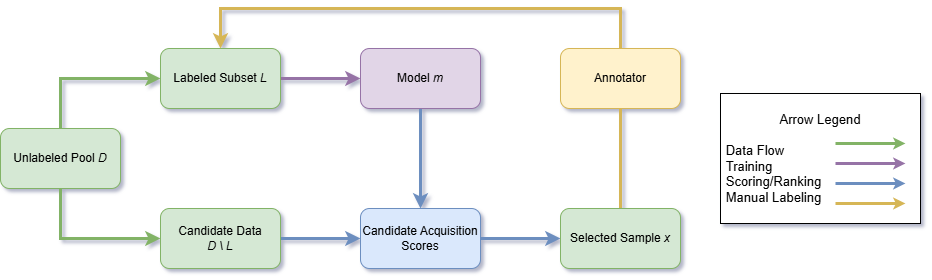}}
\caption{An illustrative diagram for Active Learning.}
\label{fig:active}
\end{figure*}

\begin{figure*}[h]
\centerline{\includegraphics[scale=0.4]{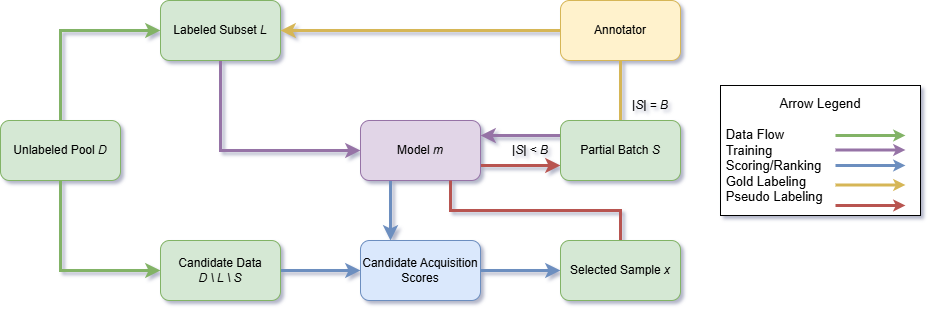}}
\caption{An illustrative diagram for ParBaLS.}
\label{fig:parbals}
\end{figure*}

\begin{figure*}[h]
\centerline{\includegraphics[scale=0.4]{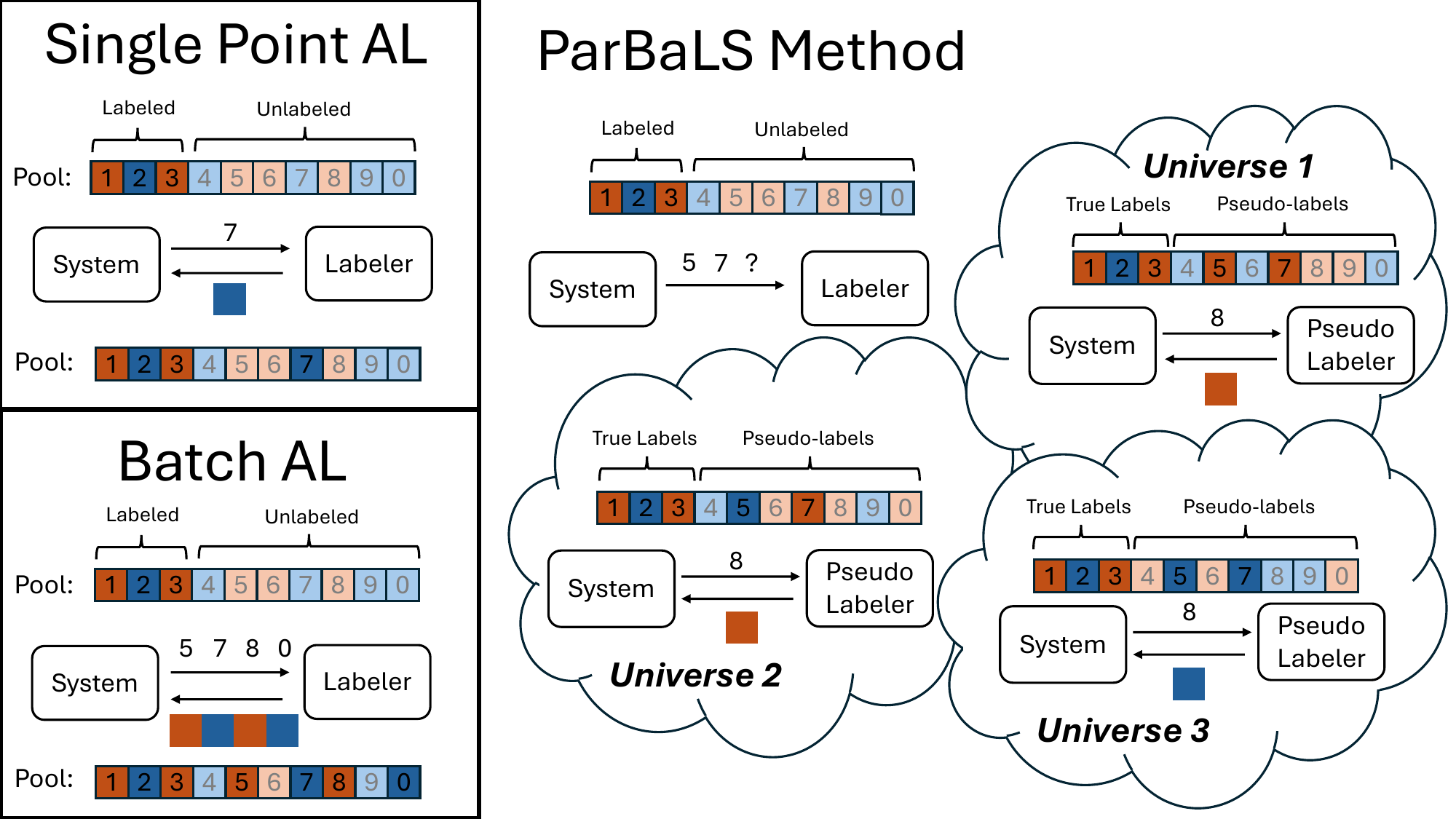}}
\caption{An illustration of the proposed method, ParBaLS.}
\label{fig:example}
\end{figure*}

As mentioned in Section \ref{sec:active}, we outline the high-level structure of Active Learning in Algorithm \ref{alg:al} and its workflow in Figure \ref{fig:active}, as well as a ParBaLS workflow in Figure \ref{fig:parbals}.

Also, as mentioned in Section \ref{sec:parbals}, we provide an illustrative example of ParBaLS in Figure \ref{fig:example}. Each square denotes a datapoint: the number is the index, and the color represents the label. A dark color means the sample is labeled (by the labeler or the pseudo-labeler). After the AL system selects the sample(s) for labeling, it sends their indices (e.g. \{7\} for Single Point AL or \{5, 7, 8, 0\} for Batch AL) to the labeler, who returns the true labels of them (e.g. \{blue\} for Single Point AL or \{red, blue, red, blue\} for Batch AL). In ParBaLS, we focus on Batch AL and reduce it to Single Point AL. In the figure, we have already committed to a partial batch of \{5, 7\}. While we don't know the true labels, we can run single point AL in alternative universes (Universe 1, 2, and 3), where we use Monte Carlo Sampling on the posterior model parameters for pseudo-labels. Within each universe, we train a model on \{1, 2, 3, 5, 7\}. We can then average the active learning acquisition scores across universes to, for example, choose 8 as the next point to label. 8 is added to the partial batch, and each universe's model is updated with the universe's pseudo-label for 8. This process continues until the partial batch is complete (with $B$ datapoints), and the AL system will send the indices of the selected batch to the labeler, as shown in Batch AL.

\begin{algorithm}[tb]
\caption{Active Learning}
\label{alg:al}
\renewcommand{\algorithmicrequire}{\textbf{Input:}}
\renewcommand{\algorithmicensure}{\textbf{Output:}}
\begin{algorithmic}[1]
\REQUIRE active learning algorithm $\mathcal{A}$;\\unlabeled pool dataset $D$;\\unlabeled validation set $V$;\\initial labeled data $L_0$ (e.g., randomly sampled) with labels $y_{L_0}$;\\model training procedure $\mathrm{Train}$;\\number of active learning iterations $T$;\\budget for each iteration $B$

\STATE Train $M_0 = \mathrm{Train}(L_0,y_{L_0})$
\FOR{iteration $t \leftarrow 1$ {\bfseries to} $T$}
\STATE $S_t \leftarrow \mathcal{A}(D, V, L_{t-1}, y_{L_{t-1}}, M_{t-1}, B)$, where $S_t \subset D$ and $|S_t| = B$
\STATE $D \leftarrow D \setminus S_t$
\STATE Acquire labels $y_{S_t}$ for $S_t$
\STATE $L_t \leftarrow L_{t-1} \cup S_t$
\STATE $y_{L_t} \leftarrow y_{L_{t-1}} \cup y_{S_t}$
\STATE Train $M_t= \mathrm{Train}(L_t, y_{L_t})$
\ENDFOR
\STATE \textbf{return} $M_T$, $L_T$, and $y_{L_T}$
\end{algorithmic}
\end{algorithm}

\section{Theoretical Derivations}\label{opt}

\subsection{Notations}

In the following derivations, we use upper-case letters for random variables, sets, and matrices, and lower-case letters for vectors and scalars.

Define $\Delta_S$ as the probability simplex over a finite set $S$. If $S$ has $n$ elements, $\Delta_S = \{p \in \mathbb{R}^n: p_i \geq 0, \sum_i p_i = 1\}$. Define $\Delta_S^k$ as the set of $k$-tuples of the probability simplex, $\Delta_S^k = \underbrace{ \Delta_S \times \dots \times \Delta_S }_{k \text{ times}}$.

Define the entropy, conditional entropy, expected conditional entropy, and mutual information as,
\begin{align}
    H(X) &= - \sum_{x} \Pr(X=x) \ln \Pr(X=x) \\
    H(X|Y=y) &= - \sum_{x} \Pr(X=x|Y=y) \ln \Pr(X=x|Y=y) \\
    H(X|Y=y,Z=z) &= - \sum_{x} \Pr(X=x|Y=y,Z=z) \ln \Pr(X=x|Y=y,Z=z) \\
    H(X|Y,Z=z) &= \sum_{y} \Pr(Y=y|Z=z) H(X|Y=y,Z=z) \\
    I(X;Y) &= H(X) - H(X|Y) \\
    I(X;Y | Z=z) &= H(X|Z=z) - H(X|Y,Z=z)
\end{align}

We use $Y_x$ to refer to the random variable of the label of a point $x$. We use $Y_S$ to refer to the random vector of labels of the points $S \subset \mathcal{X}$.

\subsection{Proof of Propositions \ref{prop:epig-derive} and \ref{prop:parbals}}
\label{app:derive-proof}

First, we prove a convenient Lemma,

\begin{lemma}
\label{lem:log-loss}
   For a random variable $A$ that takes values in a finite set $\mathcal{A}$,
   \begin{align}
       \min_{p \in \Delta_\mathcal{A}} \sum_{a \in \mathcal{A}} \Pr(A=a) \left[-\ln(p[a])\right] = H(A)
   \end{align}
\end{lemma}
\begin{proof}
    Since $-\ln(p[a])$ is convex in $p$ for any $a$, $\sum_{a \in \mathcal{A}} \Pr(A=a) \left[-\ln(p[a])\right]$ is also convex in $p$.

    At the optimal $p$, there is a Lagrange multiplier $\lambda$ corresponding to the constraint $\sum_{a \in \mathcal{A}} p[a] = 1$ such that for $a \in \mathcal{A}$,
    \begin{align}
        -\Pr(A=a) \frac{1}{p[a]} = \lambda \cdot 1
    \end{align}
    
    Solving for $p[a]$ and setting $\lambda=-1$ to satisfy the constraint, we find that the optimal $p[a] = \Pr(A=a)$. 

    Thus,
    \begin{align}
        \min_{p \in \Delta_\mathcal{A}} \sum_{a \in \mathcal{A}} \Pr(A=a) \left[-\ln(p[a])\right] &= \sum_{a \in \mathcal{A}} \Pr(A=a) \left[-\ln(\Pr(A=a))\right] \\
        &=H(A)
    \end{align}
\end{proof}

\begin{proposition-repeat}[\ref{prop:epig-derive}]
If $\ell(y,p) = - \ln p[y]$, then there is a constant $c$ that doesn't depend on $\hat{x}$ such that 
\begin{align}
    \mathrm{Cost}_{L,V,y_L}(\hat{x}) =  c - \frac{1}{|V|} \sum_{x \in V} I(Y_x; Y_{\hat{x}}|Y_L = y_L)
\end{align}
\end{proposition-repeat}
\begin{proof}
Recall that the expected cost (see Equation~\ref{eq:bdt-nonbatch}) is,
\begin{align*}
    \mathrm{Cost}_{L,V,y_L}(\hat{x}) = \mathbb{E}_{\hat{y} \sim Y_{\hat{x}}|Y_L = y_L}\left[ \min_{P \in \Delta_\mathcal{Y}^{|V|}} \mathbb{E}_{y_V \sim Y_V | Y_{\hat{x}}=\hat{y},Y_L = y_L}\left[\frac{1}{|V|} \sum_{x \in V} \ell(y_x,P_x)\right]\right]
\end{align*}

Focusing on the sub-expression, 
\begin{align}
    \min_{P \in \Delta_\mathcal{Y}^{|V|}} &\mathbb{E}_{y_V \sim Y_V | Y_{\hat{x}}=\hat{y},Y_L = y_L}\left[\frac{1}{|V|} \sum_{x \in V} \ell(y_x,P_x)\right] \\
    &= \min_{P \in \Delta_\mathcal{Y}^{|V|}} \frac{1}{|V|} \sum_{x \in V} \mathbb{E}_{y_x \sim Y_x | Y_{\hat{x}}=\hat{y},Y_L = y_L}\left[ \ell(y_x,P_x)\right] \\
    &=\frac{1}{|V|} \sum_{x \in V} \min_{p \in \Delta_\mathcal{Y}} \mathbb{E}_{y_x \sim Y_x | Y_{\hat{x}}=\hat{y},Y_L = y_L}\left[ \ell(y_x,p)\right] \\
    &=\frac{1}{|V|} \sum_{x \in V} \min_{p \in \Delta_\mathcal{Y}} \mathbb{E}_{y_x \sim Y_x | Y_{\hat{x}}=\hat{y},Y_L = y_L}\left[ -\ln p[y_x]\right] \\
    &= \frac{1}{|V|} \sum_{x \in V} H(Y_x | Y_{\hat{x}}=\hat{y},Y_L = y_L)
\end{align}

where the last line follows from  Lemma~\ref{lem:log-loss}. Then,

\begin{align}
    \mathrm{Cost}_{L,V,y_L}(\hat{x}) &= \mathbb{E}_{\hat{y} \sim Y_{\hat{x}}|Y_L=y_L}\left[P\right] \\
    &= \mathbb{E}_{\hat{y} \sim Y_{\hat{x}}|Y_L=y_L} \left[ \frac{1}{|V|} \sum_{x \in V} H(Y_x | Y_{\hat{x}}=\hat{y},Y_L = y_L) \right] \\
    &= \frac{1}{|V|} \sum_{x \in V} H(Y_x|Y_{\hat{x}},Y_L=y_L)
\end{align}

With $c=\frac{1}{|V|} \sum_{x \in V} H(Y_x|Y_L=y_L)$,

\begin{align}
    \mathrm{Cost}_{L,V,y_L}(\hat{x}) &= c - \frac{1}{|V|} \sum_{x \in V} H(Y_x|Y_L=y_L) + \frac{1}{|V|} \sum_{x \in V} H(Y_x|Y_{\hat{x}},Y_L=y_L) \\
    &= c - \frac{1}{|V|} \sum_{x \in V} I(Y_x; Y_{\hat{x}}|Y_L=y_L)
\end{align}
\end{proof}

The proof of Proposition~\ref{prop:parbals} is very similar.

\begin{proposition-repeat}[\ref{prop:parbals}]
If $\ell(y,p) = - \ln p[y]$, then there is a constant $c$ that doesn't depend on $\hat{x}$ such that 
\begin{align}
    \mathrm{Cost}_{L,S,V,y_L}(\hat{x}) =  c - \mathbb{E}_{y_S \sim Y_S|Y_L=y_L} \left[ \frac{1}{|V|} \sum_{x \in V} I(Y_x; Y_{\hat{x}}| Y_S = y_S,Y_L=y_L) \right]
\end{align}
\end{proposition-repeat}
\begin{proof}
    Recall from Equation~\ref{eq:parbals-cost},
\begin{align}
   \mathrm{Cost}_{L,S,V,y_L}(\hat{x}) =\mathbb{E}_{\hat{y},y_S \sim Y_{\hat{x}},Y_S|Y_L=y_L} \left[ \min_{P \in \Delta_\mathcal{Y}^{|V|}} \mathbb{E}_{y_V \sim Y_V | Y_{\hat{x}}=\hat{y},Y_S=y_S,Y_L=y_L}\left[\frac{1}{|V|} \sum_{x \in V} \ell(y_x,P_x)\right]\right]
\end{align}

Define 

\begin{align}
\min_{P \in \Delta_\mathcal{Y}^{|V|}} &\mathbb{E}_{y_V \sim Y_V | Y_{\hat{x}}=\hat{y},Y_S=y_S,Y_L=y_L}\left[\frac{1}{|V|} \sum_{x \in V} \ell(y_x,P_x)\right] \\
&= \min_{P \in \Delta_\mathcal{Y}^{|V|}} \frac{1}{|V|} \sum_{x \in V} \mathbb{E}_{y_V \sim Y_V | Y_{\hat{x}}=\hat{y},Y_S=y_S,Y_L=y_L}\left[ \ell(y_x,P_x)\right] \\
&= \frac{1}{|V|} \sum_{x \in V} \min_{p \in \Delta_\mathcal{Y}}  \mathbb{E}_{y_V \sim Y_V | Y_{\hat{x}}=\hat{y},Y_S=y_S,Y_L=y_L}\left[ \ell(y_x,p)\right] \\
&= \frac{1}{|V|} \sum_{x \in V} \min_{p \in \Delta_\mathcal{Y}}  \mathbb{E}_{y_V \sim Y_V | Y_{\hat{x}}=\hat{y},Y_S=y_S,Y_L=y_L}\left[ -\ln p[y_x]\right] \\
&= \frac{1}{|V|} \sum_{x \in V} H(Y_x | Y_{\hat{x}}=\hat{y},Y_S=y_S,Y_L = y_L)
\end{align}

\begin{align}
    \mathrm{Cost}_{L,S,V,y_L}(\hat{x}) &= \mathbb{E}_{\hat{y},y_S \sim Y_{\hat{x}},Y_S|Y_L=y_L}\left[P\right] \\
    &= \mathbb{E}_{\hat{y},y_S \sim Y_{\hat{x}},Y_S|Y_L=y_L} \left[ \frac{1}{|V|} \sum_{x \in V} H(Y_x | Y_{\hat{x}}=\hat{y},Y_S=y_S,Y_L = y_L) \right] \\
    &= \frac{1}{|V|} \sum_{x \in V} \mathbb{E}_{y_S \sim Y_S | Y_L = y_L} \left[ H(Y_x|Y_{\hat{x}},Y_S=y_S,Y_L=y_L) \right]
\end{align}

With $c=\frac{1}{|V|} \sum_{x \in V} \mathbb{E}_{y_S \sim Y_S | Y_L = y_L} \left[ H(Y_x|Y_S=y_S,Y_L=y_L) \right]$,

\begin{align}
    \mathrm{Cost}_{L,S,V,y_L}(\hat{x}) &= c - \frac{1}{|V|} \sum_{x \in V} \mathbb{E}_{y_S \sim Y_S | Y_L = y_L} \left[ H(Y_x|Y_S=y_S,Y_L=y_L) \right] \\ &+ \frac{1}{|V|}\sum_{x \in V} \mathbb{E}_{y_S \sim Y_S | Y_L = y_L} \left[ H(Y_x|Y_{\hat{x}},Y_S=y_S,Y_L=y_L)\right]\\
    &= c - \frac{1}{|V|}\sum_{x \in V} \mathbb{E}_{y_S \sim Y_S | Y_L = y_L} \left[ I(Y_x; Y_{\hat{x}}|Y_S = y_S, Y_L=y_L) \right]
\end{align}
\end{proof}

\subsection{Statement and Proof of Proposition \ref{prop:mc-approx}}\label{app:proof-mc-approx}

\begin{proposition}
\label{prop:mc-approx}
 Let $x^\star \in \argmin_{x \in D} \mathrm{Cost}_{L,S,V,y_L}(x)$. For any $\delta\in(0,1)$, with probability $1-\delta$ over sampling $\{y^{(i)}\}_{i=1}^m$,
    \begin{align*}
         \mathrm{Cost}_{L,S,V,y_L}(\mathrm{ParBaLS}(D,L,S,V,y_L,\{y^{(i)}\}_{i=1}^m)) - \mathrm{Cost}_{L,S,V,y_L}(x^\star) \leq 2 \ln( |\mathcal{Y}| ) \sqrt{\frac{\ln(|D|) + \ln(1/\delta)}{2m}}
    \end{align*}
\end{proposition}

\begin{proof}
For a fixed $V$, $L$, $S$, and $y_L$, define $f(\hat{x},y_S) = \frac{1}{|V|} \sum_{x \in V} I(Y_x; Y_{\hat{x}}| Y_S = y_S,Y_L=y_L)$. Note that $0 \leq f(\hat{x},y_S) \leq \ln |\mathcal{Y}|$. 

Let $\hat{x}(\{y^{(i)}\}_{i=1}^m) = \mathrm{ParBaLS}(D,L,S,V,y_L,\{y^{(i)}\}_{i=1}^m)$. Using Proposition~\ref{prop:parbals},

\begin{align}
    x^\star &\in \argmax_{\hat{x} \in D} \mathbb{E}_{y_S \sim Y_S|Y_L = y_L} \left[ \frac{1}{|V|} \sum_{x \in V} I(Y_x; Y_{\hat{x}}| Y_S = y_S,Y_L=y_L) \right] = \argmax_{\hat{x} \in D} \mathbb{E}_{y_S \sim Y_S|Y_L=y_L}[f(\hat{x},y_S)] \\
    \hat{x}(\{y^{(i)}\}_{i=1}^m) &\in \argmax_{\hat{x} \in D} \frac{1}{m |V|} \sum_{i=1}^m \sum_{x \in V} I(Y_x; Y_{\hat{x}}| Y_S = y^{(i)}_S,Y_L=y_L) = \argmax_{\hat{x} \in D} \frac{1}{m} \sum_{i=1}^m f(\hat{x}, y_S^{(i)})
\end{align}

From Hoeffding's inequality, for any fixed $x \in D \setminus \{x^\star\}$ and any $\epsilon>0$,
\begin{align}
    \Pr\left(\frac{1}{m} \sum_{i=1}^m f(x,y_S^{(i)}) - \mathbb{E}_{y_S \sim Y_S|Y_L=y_L}[f(x,y_S)] \geq\epsilon\right) \leq \exp\left( - \frac{2m\epsilon^2}{\ln^2(|\mathcal{Y}|)} \right)
\end{align}

Also,
\begin{align}
    \Pr\left(\frac{1}{m} \sum_{i=1}^m f(x^\star,y_S^{(i)}) - \mathbb{E}_{y_S \sim Y_S|Y_L=y_L}[f(x^\star,y_S)] \leq -\epsilon\right) \leq \exp\left( - \frac{2m\epsilon^2}{\ln^2(|\mathcal{Y}|)} \right)
\end{align}

Union bounding over all elements of $D$, we have that with probability at least $1-|D| \exp\left( - \frac{2m\epsilon^2}{\ln^2(|\mathcal{Y}|)} \right)$,
\begin{align}
    \forall x \neq x^\star: &\frac{1}{m} \sum_{i=1}^m f(x,y_S^{(i)}) - \mathbb{E}_{y_S \sim Y_S|Y_L=y_L}[f(x,y_S)] \leq \epsilon \\
    &\frac{1}{m} \sum_{i=1}^m f(x^\star,y_S^{(i)}) - \mathbb{E}_{y_S \sim Y_S|Y_L=y_L}[f(x^\star,y_S)] \geq -\epsilon
\end{align}

Under such an event,

\begin{align}
    \mathbb{E}_{y_S \sim Y_S|Y_L=y_L}[f(x^\star,y_S)] - &\mathbb{E}_{y_S \sim Y_S|Y_L = y_L}[f(\hat{x}(\{y^{(i)}\}_{i=1}^m),y_S)] = \\
    &= \left[\mathbb{E}_{y_S \sim Y_S|Y_L=y_L}[f(x^\star,y_S)] -  \frac{1}{m} \sum_{i=1}^m f(x^\star,y_S^{(i)})\right] + \\
    &+ \left[\frac{1}{m} \sum_{i=1}^m f(x^\star,y_S^{(i)}) -  \frac{1}{m} \sum_{i=1}^m f(\hat{x}(\{y^{(i)}\}_{i=1}^m),y_S^{(i)})\right] + \\
    &+ \left[\frac{1}{m} \sum_{i=1}^m f(\hat{x}(\{y^{(i)}\}_{i=1}^m),y_S^{(i)}) - \mathbb{E}_{y_S \sim Y_S|Y_L = y_L}[f(\hat{x}(\{y^{(i)}\}_{i=1}^m),y_S)] \right] \\
    &\leq \epsilon + 0 + \epsilon \\
    &= 2\epsilon
\end{align}

From Proposition~\ref{prop:parbals}, we know that for any $x$ and $x'$ (in particular $x=x^\star$ and $x'=\hat{x}(\{y^{(i)}\}_{i=1}^m)$),
\begin{align}
    \mathbb{E}_{y_S \sim Y_S|Y_L=y_L}[f(x,y_S)] - &\mathbb{E}_{y_S \sim Y_S|Y_L = y_L}[f(x',y_S)] = \mathrm{Cost}_{L,S,V,y_L}(x') - \mathrm{Cost}_{L,S,V,y_L}(x)
\end{align}

Setting $\delta$ to be the failure probability and solving for $\epsilon$,

\begin{align}
    \delta &= |D| \exp\left( - \frac{2m\epsilon^2}{\ln^2(|\mathcal{Y}|)} \right) \\
    \epsilon &= \sqrt{\frac{\ln^2(|\mathcal{Y}|) \ln(|D|/\delta)}{2m}} \\
    &= \ln(|\mathcal{Y}|) \sqrt{\frac{\ln(|D|) + \ln(1/\delta)}{2m}}  
\end{align}

The previous three statements combine to give the result.
\end{proof}

\section{Comparison to BatchBALD}\label{sec:batch}

The key idea of BatchBALD \citep{kirsch2019batchbald} is to greedily build a batch using the joint score over the batch. If this were ported directly to EPIG, it would be:

\begin{equation}
    \begin{aligned}\label{eq:batchepig}
        &\text{BatchEPIG}(D,L,S,V,y_L) \in \arg\max_{\hat{x} \in D} \sum_{x \in V} I(Y_x; Y_{\hat{x}}, Y_S| Y_L=y_L)
    \end{aligned}
\end{equation}

This has two computational issues. First, exactly computing $I(Y_x; Y_{\hat{x}}, Y_S| Y_L=y_L)$ requires summing over all $|\mathcal{Y}|^{|S|+2}$ possible values of $(Y_x, Y_{\hat{x}}, Y_S)$. However, this computational issue can be addressed by Monte Carlo importance sampling, as described in Appendix C of \citet{kirsch2019batchbald}, though the sampling distribution is left unspecified. 

The more critical computational issue is the large number of parameter posterior samples required to accurately estimate large joint probabilities, like $\Pr(Y_S = y_S)$ for large $S$. As a simple generic example, examine binary classification in $d$ dimensions, where the parameter posterior (based on the labeled data) is an isotropic Gaussian, $\theta \sim \mathcal{N}(0, \sigma^2 I_d)$, and the partial batch contains a set of $B$ orthonormal vectors, for example, the first $B$ basis vectors. From symmetry, the distribution of $Y_S$ is uniform over $\{-1,1\}^B$, i.e., $\Pr(Y_S = y_S) = 1/2^B$, for any $y_S$. However, for a posterior sample $\theta^{(j)}$, $\Pr(Y_S = y_S | \theta^{(j)})$ will be exponentially decaying in $\|\theta^{(j)}\|$ (large if $\sigma^2$ is reasonably large) unless $\theta^{(j)}$ is in the orthant corresponding to $y_S$, $\{\theta: \forall i \in [B], y_i\theta_i \geq 0\}$, which occurs with $1/2^B$ probability. Thus, estimating $\Pr(Y_S)$ using Monte Carlo parameter posterior samples requires the number of samples to be exponential in the size of $|S|=B$.

ParBaLS avoids this issue by only requiring the computation of joint probabilities for 2 variables, $Y_x$ (for $x \in V$) and $Y_{\hat{x}}$. This key property is enabled by retraining on the sampled pseudo-labels.

\section{Additional Experimental Details}\label{sec:results}

\subsection{Detailed Experimental Setup}\label{sec:setup}

\begin{table}
\centering
\begin{tabular}{lcccc}
\hline
Dataset & Encoder & Before PCA & After PCA \\ \hline
CIFAR-10 & DINOv2-ViT-S/14 & 384 & 348 \\
iWildCam & CLIP-ViT-B/32 & 512 & 305 \\
fMoW & CLIP-ViT-B/32 & 512 & 342 \\
AG News & bert-base-uncased & 768 & 210 \\
Yelp Review Full & bert-base-uncased & 768 & 168 \\ 
Civil Comments & bert-base-uncased & 768 & 187 \\ \hline
\end{tabular}
\caption{Encoders used for each dataset and the embedding dimensions before and after 99\% PCA.}
\label{tab:dim}
\end{table}

\paragraph{EPIG.} Following \cite{mussmann2022active}, for EPIG \citep{kirsch2021test, mussmann2022active, smith2023prediction}, we randomly subsample a subset of the candidate pool and validation set during each iteration, whose sizes are 10000 and 1000, respectively.

\paragraph{GLISTER.} For GLISTER \citep{killamsetty2021glister}, the original paper uses 10\% of the initial labeled training set as the labeled validation set when there is no separate validation set. However, we use 20\% of the initial labeled set as the labeled validation set, as in our low-budget initialization settings, 10\% can be too small, e.g., only 2 data points for validation when the initial labeled set only has 20 samples.

\paragraph{BatchBALD.} BatchBALD \citep{kirsch2019batchbald} is too slow and can run out of extensive memory with the larger batch sizes, so we only include it in the third budget setting ($B=10$ rather than $B=20$).

\paragraph{Feature preprocessing.} To avoid numeric issues, we apply PCA to reduce the embedding space while retaining 99\% of the variance. The details of the encoders and dimensions before and after PCA for each dataset are shown in Table \ref{tab:dim}. For tabular datasets, we directly apply Bayesian Logistic Regression after feature preprocessing (apply quantile binning with 10 bins to numerical features, then apply one-hot encoding to all features). Both PCA and quantile are fitted using the features from the pool dataset and fixed throughout the iterations.

\paragraph{Bayesian Logistic Regression.} If $\mathcal{X} = \mathbb{R}^d$ and there are $c$ classes, a logistic regression predictor is defined by weights $W \in \mathbb{R}^{c \times d}$ and biases $b \in \mathbb{R}^c$, trained by minimizing logistic loss. Bayesian Logistic Regression (BLR) instead places a Gaussian prior over the parameters and performs Bayesian model averaging over the posterior. We approximate inference with Monte Carlo sampling using the NUTS sampler \citep{hoffman2014no} implemented in PyMC \citep{AbrilPla2023PyMCAM}, the default sampler in PyMC. We use Gaussian distribution $\mathcal{N}(\mu=0,\sigma=1)$ for the prior following the default setup in PyMC. We use $k = 400$ posterior parameter samples in all experiments, which is chosen based on a balance of effectiveness and efficiency in preliminary random selection experiments.

\paragraph{Environment, metrics, and statistical interpretation.} We conduct our experiments with the NVIDIA RTX 6000 GPU on a Slurm-based cluster \citep{10.1007/10968987_3}, using the LabelBench package \citep{zhang2024labelbench}, a well-established framework to benchmark label-efficient learning, including active learning. For each setting, we conduct 10 runs with different seeds to gauge the variability of our results. Following EPIG evaluations \citep{smith2023prediction}, we mainly report test accuracy as the evaluation metric. In addition, we measure the test negative log likelihood (NLL), which is aligned with the loss used to derive our acquisition objective. We report the average performance over the 10 runs and show the 95\% confidence interval from the Student's $t$ distribution\footnote{\url{https://docs.scipy.org/doc/scipy/reference/generated/scipy.stats.t.html}} using ``$\pm$''. We consider a method to be ``Highest Mean (HM)'' (bolded and underlined in our presented tables) if it has the highest mean performance in a given data setting. We additionally consider a method to be ``Ties with HM'' (bolded in our presented tables) if the p-value from Welch's two-sample $t$-test\footnote{\url{https://docs.scipy.org/doc/scipy/reference/generated/scipy.stats.mstats.ttest_ind.html}} between its per-run values and those of the highest-mean (HM) method is greater than or equal to $0.05$.

\subsection{Detailed Main Experiment Results}\label{sec:tables}

\begin{figure*}[t]
\centerline{\includegraphics[width=\textwidth]{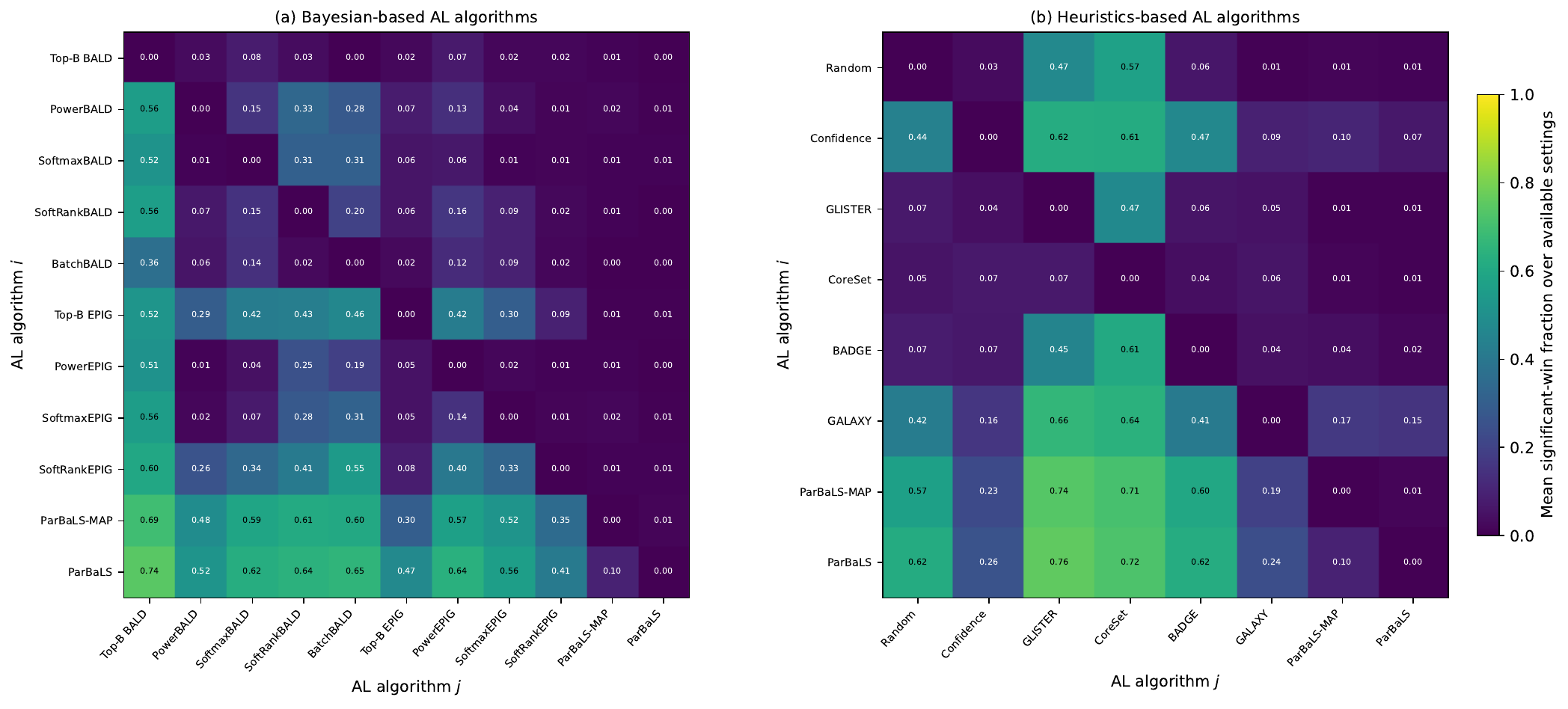}}
\caption{Pairwise statistical-significance win matrices across all acquisition rounds.}
\label{fig:win-matrices}
\end{figure*}

Following the win-matrix analysis of \citet{gupte2024revisiting}, we also compare every pair of strategies at every iteration. For a given data setting and iteration, we form paired accuracy differences over the ten matched seeds and count a directed win when the paired $t$ statistic is significantly positive at $p<0.05$. Within each setting, we compute the fraction of acquisition rounds won, then average this fraction over all settings available to that strategy pair. Every cell in Figure \ref{fig:win-matrices} therefore lies on the same $[0,1]$ scale, including comparisons involving BatchBALD, which is available only for the 11 settings with $B=10$. Larger values indicate that row strategy $i$ significantly outperforms column strategy $j$ more consistently throughout the learning curve. ParBaLS has the strongest net-win score in both comparisons, with ParBaLS-MAP as its closest competitor.

We then provide the paired global comparison of ParBaLS with Top-$B$ EPIG in Table \ref{tab:global-paired}, which gives explicit paired global comparisons across all 11 datasets, where we pair runs by seed, average seed-level differences within each data setting, and then average repeated budget settings within each dataset. We evaluate both accuracy and the NLL used in the Bayesian decision-theoretic derivation. We report the mean dataset-level effect and a 95\% Student $t$ interval across datasets. The positive mean improvement and interval quantify the significance of the results, which are shown more comprehensively in Figure \ref{fig:win-matrices}.

\begin{table}[t]
\centering
\begin{tabular}{lcc}
\hline
Metric & Mean improvement & 95\% interval \\ \hline
Final test accuracy & $+1.54$ pp & $[+0.46,+2.62]$ \\
Iteration-averaged test accuracy & $+1.25$ pp & $[+0.60,+1.91]$ \\
Final test NLL reduction & 0.0112 & $[0.0052,0.0172]$ \\
Iteration-averaged test NLL reduction & 0.0086 & $[0.0057,0.0115]$ \\ \hline
\end{tabular}
\caption{Paired global comparison of ParBaLS with Top-$B$ EPIG across 10 seeds, 11 datasets, and 3 budget settings.}
\label{tab:global-paired}
\end{table}

Then, we provide the full tables for our experiment results shown in Section \ref{exp} with a roadmap in Table \ref{tab:main_results_guide}.

\begin{table*}
\centering
\begin{tabular}{l|c|c|c}
\hline
Initial Budget & Acquisition Budget
& Bayesian Methods & Heuristic Methods \\ \hline

100
& $10\times20$
& Table~\ref{tab:bs100_bayesian_ttest}
& Table~\ref{tab:bs100_heuristic_ttest} \\

20
& $10\times20$
& Table~\ref{tab:bs20_bayesian_ttest}
& Table~\ref{tab:bs20_heuristic_ttest} \\

100
& $10\times10$
& Table~\ref{tab:bs10_bayesian_ttest}
& Table~\ref{tab:bs10_heuristic_ttest} \\ \hline

\end{tabular}
\caption{Guide to the main active learning result tables. ``Initial Budget''
denotes the number of randomly labeled samples used for initialization.
``Acquisition Budget'' gives the number of active learning iterations and the
per-iteration labeling budget.}
\label{tab:main_results_guide}
\end{table*}

\subsection{Ablation Studies on Various Numbers of Sampled Pseudo-labels}\label{ablations}

In this section, we focus on Airline Passenger Satisfaction, Credit Card Fraud, and CIFAR-10, where each setting is conducted with 5 different seeds. We highlight results whose configurations are used in our main experiments in \main{blue font}.

For the standard budget settings, along with ParBaLS-MAP EPIG, we provide additional experimental results using three different numbers of universes for ParBaLS EPIG ($m=5,10,20$) in Table \ref{tab:parbals100blr10by20}, where $m=10$ is the default setting in the main paper.

\begin{table*}
\centering
\begin{tabular}{l|ccc}\hline
Datasets & Airline Passenger & Credit Card & \\
Algorithm & Satisfaction & Fraud & CIFAR-10 \\ \hline
ParBaLS-MAP EPIG & \main{\textbf{89.34$\pm$0.46}} & \main{\textbf{93.45$\pm$0.29}} & \main{\textbf{93.40$\pm$1.19}} \\
ParBaLS EPIG ($m=5$) & \textbf{89.45$\pm$0.58} & \textbf{93.37$\pm$0.13} & \textbf{93.80$\pm$1.18} \\
ParBaLS EPIG ($m=10$) & \main{\textbf{89.35$\pm$0.60}} & \main{\textbf{93.45$\pm$0.22}} & \main{\underline{\textbf{94.90$\pm$0.54}}} \\
ParBaLS EPIG ($m=20$) & \underline{\textbf{89.70$\pm$0.55}} & \underline{\textbf{93.49$\pm$0.20}} & \textbf{94.17$\pm$0.72} \\ \hline
\end{tabular}
\caption{Final test accuracy of different variants of ParBaLS with Bayesian Logistic Regression, where each of the 10 iterations has a labeling budget of 20 samples, except for the first iteration that starts with 100 samples.}
\label{tab:parbals100blr10by20}
\end{table*}

For different values of $m$, we note that ParBaLS EPIG ($m=5$) does slightly worse than the other two $m$s. On the other hand, ParBaLS EPIG ($m=20$) does not always perform the best, and its computational complexity is twice as much as ParBaLS EPIG ($m=10$). In practice, ParBaLS EPIG ($m=10$) is the most solid choice, given its balance between performance and complexity.

\subsection{More Learning Curves}\label{sec:curves}

\begin{figure}[h]
\centerline{\includegraphics[scale=0.35]{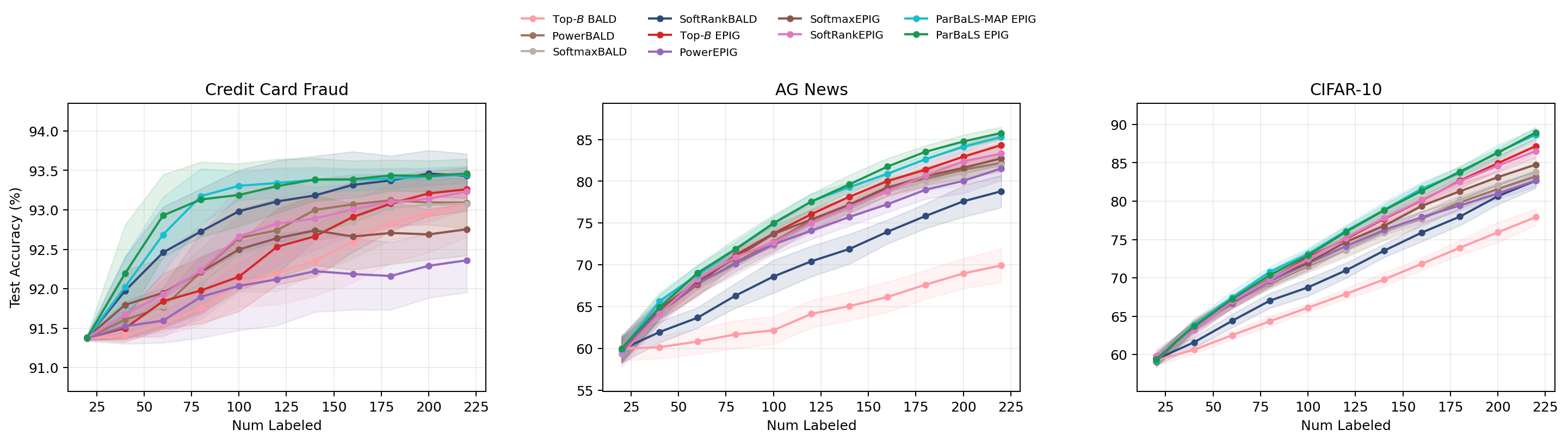}}
\caption{Learning curves of Bayesian AL algorithms with Bayesian Logistic Regression, where each of the 10 iterations has a labeling budget of 20 samples, starting with random initialization of 20 samples.}
\label{fig:bs20_bayesian}
\end{figure}

\begin{figure}[h]
\centerline{\includegraphics[scale=0.35]{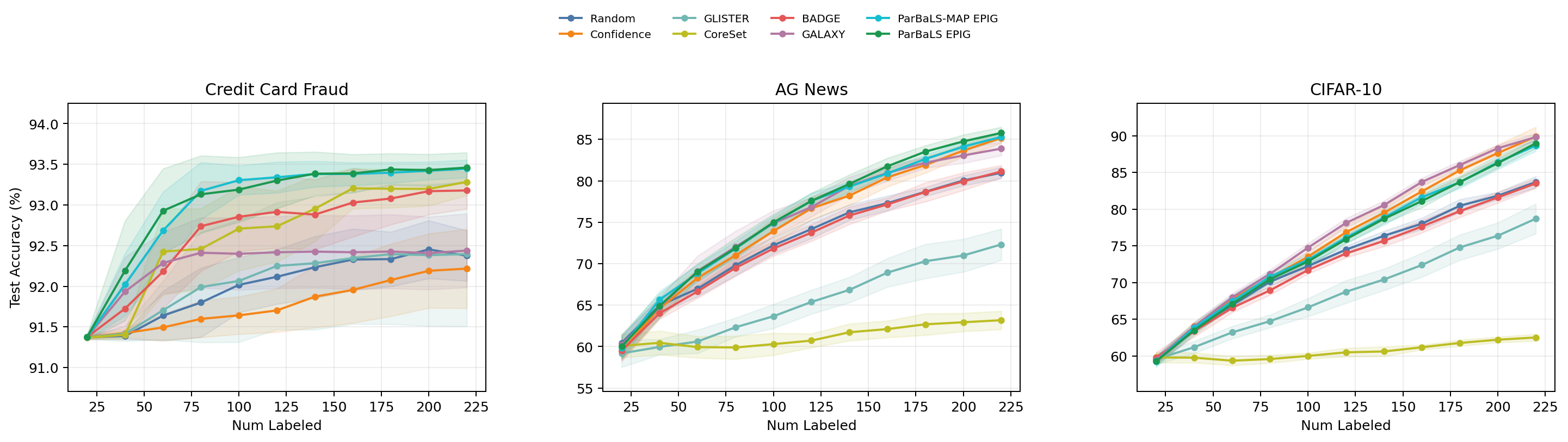}}
\caption{Learning curves of ParBaLS and heuristics baselines with Bayesian Logistic Regression, where each of the 10 iterations has a labeling budget of 20 samples, starting with random initialization of 20 samples.}
\label{fig:bs20_heuristic}
\end{figure}

\begin{figure}[h]
\centerline{\includegraphics[scale=0.35]{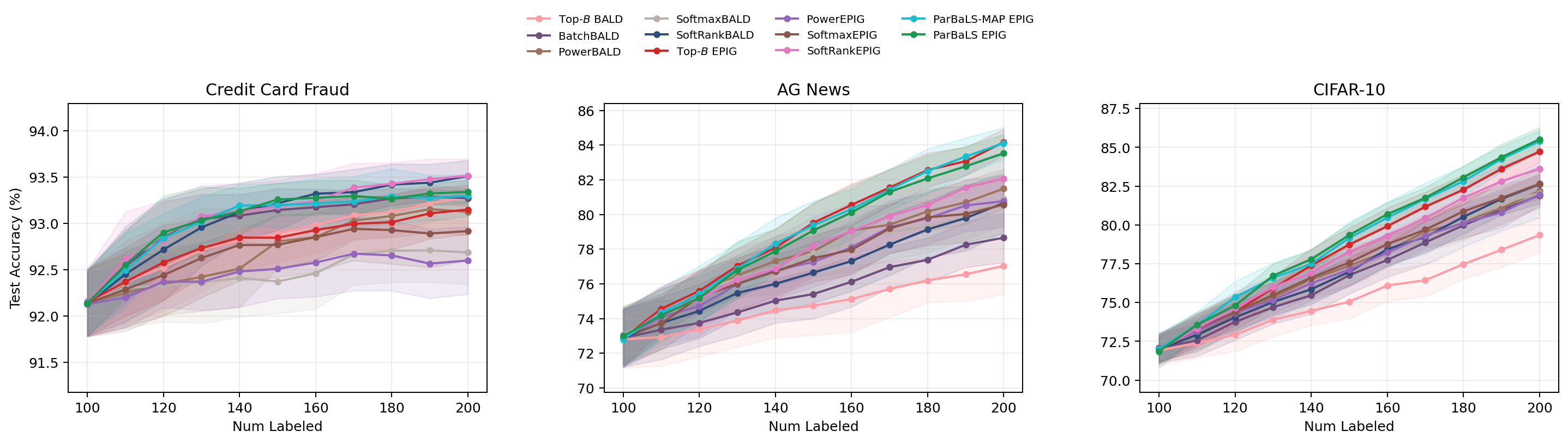}}
\caption{Learning curves of Bayesian AL algorithms with Bayesian Logistic Regression, where each of the 10 iterations has a labeling budget of 10 samples, starting with random initialization of 100 samples.}
\label{fig:bs10_bayesian}
\end{figure}

\begin{figure}[h]
\centerline{\includegraphics[scale=0.35]{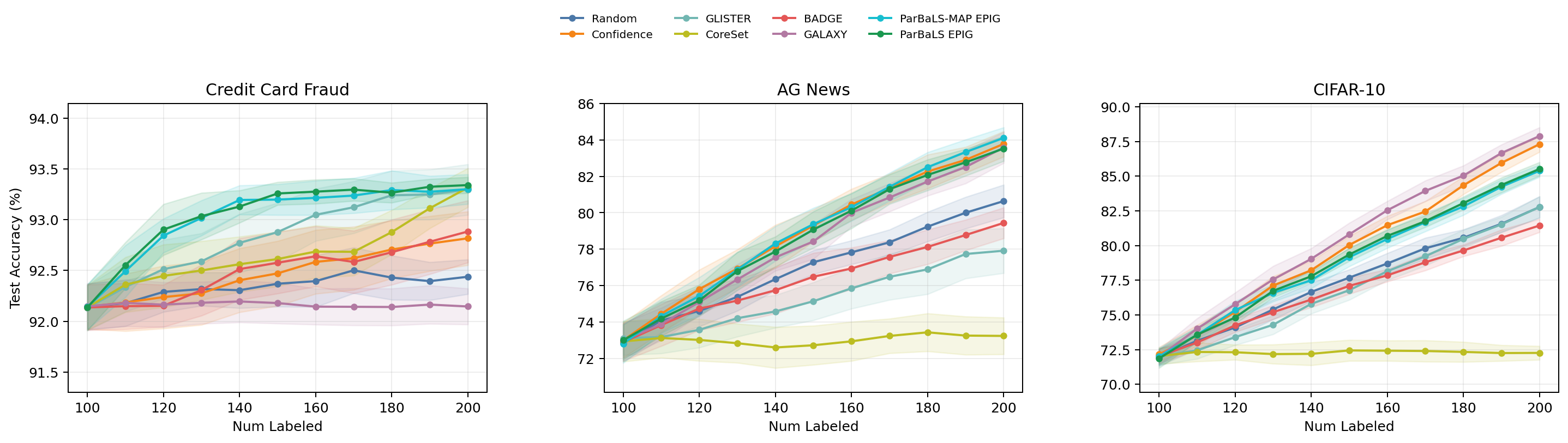}}
\caption{Learning curves of ParBaLS and heuristics baselines with Bayesian Logistic Regression, where each of the 10 iterations has a labeling budget of 10 samples, starting with random initialization of 100 samples.}
\label{fig:bs10_heuristic}
\end{figure}

Apart from the learning curves in Section \ref{exp}, we also present the full learning curves with more budget settings in Figure \ref{fig:bs20_bayesian}, Figure \ref{fig:bs20_heuristic}, Figure \ref{fig:bs10_bayesian}, and Figure \ref{fig:bs10_heuristic}.

\subsection{More Variants of One-vs-all Datasets}\label{ova_ood}

Apart from taking the first original class as the ``one'' and the rest as the ``all'', we also conduct experiments with other choices of the ``one'' for each multiclass dataset with the standard budget and the low-budget initialization, with a roadmap in Table \ref{tab:ova_results_guide}.

\begin{table*}
\centering
\begin{tabular}{l|c|c|c|c}
\hline
Dataset & Initial Budget & Acquisition Budget
& Bayesian Methods & Heuristic Methods \\ \hline

CIFAR-10
& 20
& $10\times20$
& Table~\ref{tab:cifar10_ova_bs20_bayesian_ttest}
& Table~\ref{tab:cifar10_ova_bs20_heuristic_ttest} \\

CIFAR-10
& 100
& $10\times20$
& Table~\ref{tab:cifar10_ova_bs100_bayesian_ttest}
& Table~\ref{tab:cifar10_ova_bs100_heuristic_ttest} \\

iWildCam
& 20
& $10\times20$
& Table~\ref{tab:iwildcam_ood_bs20_bayesian_ttest}
& Table~\ref{tab:iwildcam_ood_bs20_heuristic_ttest} \\

iWildCam
& 100
& $10\times20$
& Table~\ref{tab:iwildcam_ood_bs100_bayesian_ttest}
& Table~\ref{tab:iwildcam_ood_bs100_heuristic_ttest} \\

fMoW
& 20
& $10\times20$
& Table~\ref{tab:fmow_ood_bs20_bayesian_ttest}
& Table~\ref{tab:fmow_ood_bs20_heuristic_ttest} \\

fMoW
& 100
& $10\times20$
& Table~\ref{tab:fmow_ood_bs100_bayesian_ttest}
& Table~\ref{tab:fmow_ood_bs100_heuristic_ttest} \\

AG News
& 20
& $10\times20$
& Table~\ref{tab:agnews_ova_bs20_bayesian_ttest}
& Table~\ref{tab:agnews_ova_bs20_heuristic_ttest} \\

AG News
& 100
& $10\times20$
& Table~\ref{tab:agnews_ova_bs100_bayesian_ttest}
& Table~\ref{tab:agnews_ova_bs100_heuristic_ttest} \\

Yelp
& 20
& $10\times20$
& Table~\ref{tab:yelp_ova_bs20_bayesian_ttest}
& Table~\ref{tab:yelp_ova_bs20_heuristic_ttest} \\

Yelp
& 100
& $10\times20$
& Table~\ref{tab:yelp_ova_bs100_bayesian_ttest}
& Table~\ref{tab:yelp_ova_bs100_heuristic_ttest} \\ \hline

\end{tabular}
\caption{Guide to the one-vs-all active learning result tables. ``Initial Budget''
denotes the number of randomly labeled samples used for initialization.
``Acquisition Budget'' gives the number of active learning iterations and the
per-iteration labeling budget.}
\label{tab:ova_results_guide}
\end{table*}

\begin{table*}
\centering
\resizebox{\columnwidth}{!}{%
\begin{tabular}{l|ccccc|ccc|ccc}
\hline
Datasets & \multicolumn{5}{c|}{Tabular Datasets} & \multicolumn{3}{c|}{Text Datasets} & \multicolumn{3}{c}{Image Datasets} \\
Algorithm & Airline & Credit Card & Magic & Satellite & Vowels & Civil Comments & AG News & Yelp & CIFAR-10 & iWildCam & fMoW \\ \hline
Top-$B$ BALD & 87.07$\pm$0.83 & \textbf{93.41$\pm$0.18} & \underline{\textbf{68.73$\pm$2.09}} & \textbf{82.59$\pm$0.28} & \textbf{43.46$\pm$0.20} & 82.19$\pm$2.49 & 80.19$\pm$1.80 & 71.58$\pm$1.41 & 88.25$\pm$0.93 & 78.62$\pm$2.83 & 91.78$\pm$3.84 \\
PowerBALD & 88.43$\pm$0.46 & 93.18$\pm$0.20 & 65.09$\pm$1.33 & \textbf{82.69$\pm$0.37} & \textbf{43.64$\pm$0.43} & 86.67$\pm$0.89 & 85.17$\pm$1.02 & 78.51$\pm$1.14 & 90.06$\pm$0.75 & 82.93$\pm$4.62 & 91.42$\pm$1.18 \\
SoftmaxBALD & 87.42$\pm$0.74 & 92.75$\pm$0.24 & 65.11$\pm$2.36 & 82.03$\pm$0.61 & \textbf{43.23$\pm$0.68} & 85.25$\pm$0.98 & 85.29$\pm$0.75 & 78.11$\pm$1.22 & 88.90$\pm$0.79 & 81.99$\pm$4.63 & 90.44$\pm$1.51 \\
SoftRankBALD & 88.65$\pm$0.45 & \textbf{93.34$\pm$0.13} & 65.66$\pm$1.82 & 82.41$\pm$0.23 & \textbf{43.59$\pm$0.37} & 86.35$\pm$1.64 & 84.47$\pm$1.82 & 75.39$\pm$1.28 & 91.74$\pm$0.86 & 85.27$\pm$1.23 & 94.12$\pm$2.02 \\
\hline
Top-$B$ EPIG & \textbf{89.05$\pm$0.44} & \textbf{93.35$\pm$0.14} & 65.45$\pm$2.26 & \textbf{82.70$\pm$0.19} & \textbf{43.51$\pm$0.19} & \textbf{88.41$\pm$0.95} & \textbf{87.54$\pm$0.79} & 79.30$\pm$0.99 & 93.82$\pm$1.02 & 85.80$\pm$3.00 & 97.06$\pm$0.72 \\
PowerEPIG & 87.72$\pm$0.66 & 92.95$\pm$0.24 & 65.01$\pm$1.10 & 82.41$\pm$0.41 & \underline{\textbf{43.68$\pm$0.40}} & 83.92$\pm$1.15 & 84.51$\pm$1.20 & 77.50$\pm$0.76 & 88.54$\pm$0.73 & 84.12$\pm$3.38 & 90.59$\pm$0.80 \\
\hline
\end{tabular}
}
\caption{Bayesian AL comparison: $100+10\times20$. The reported number is final test accuracy averaged over 10 seeds with $\pm$ defining the $95\%$ confidence interval using Student's $t$-distribution. The highest mean accuracy is underlined, and all methods with means that are not significantly different ($p$-value greater than or equal to $0.05$) according to Welch's $t$-test are bolded.}
\label{tab:bs100_bayesian_ttest}
\end{table*}

\begin{table*}
\centering
\resizebox{\columnwidth}{!}{%
\begin{tabular}{l|ccccc|ccc|ccc}
\hline
Datasets & \multicolumn{5}{c|}{Tabular Datasets} & \multicolumn{3}{c|}{Text Datasets} & \multicolumn{3}{c}{Image Datasets} \\
Algorithm & Airline & Credit Card & Magic & Satellite & Vowels & Civil Comments & AG News & Yelp & CIFAR-10 & iWildCam & fMoW \\ \hline
Top-$B$ BALD & 84.50$\pm$2.10 & \textbf{93.07$\pm$0.42} & \textbf{67.72$\pm$4.32} & \textbf{82.52$\pm$0.21} & 41.30$\pm$0.61 & 70.83$\pm$1.87 & 69.93$\pm$2.01 & 63.60$\pm$1.22 & 77.93$\pm$1.06 & 66.73$\pm$4.06 & 75.47$\pm$2.40 \\
PowerBALD & 87.62$\pm$0.62 & 93.09$\pm$0.33 & \textbf{67.40$\pm$1.76} & \textbf{82.29$\pm$0.49} & \textbf{41.88$\pm$1.90} & 82.40$\pm$1.52 & 82.21$\pm$1.08 & \textbf{77.86$\pm$0.89} & 83.22$\pm$0.72 & \textbf{79.84$\pm$4.44} & 86.60$\pm$2.14 \\
SoftmaxBALD & 86.70$\pm$0.59 & 93.08$\pm$0.30 & \textbf{66.66$\pm$2.28} & 81.81$\pm$0.66 & 41.60$\pm$1.43 & 83.03$\pm$1.39 & 81.96$\pm$1.41 & 76.98$\pm$1.26 & 83.86$\pm$1.11 & \textbf{81.31$\pm$4.93} & 87.06$\pm$1.78 \\
SoftRankBALD & 87.88$\pm$0.46 & \textbf{93.43$\pm$0.28} & 65.01$\pm$2.20 & 81.71$\pm$0.37 & 41.84$\pm$0.74 & 82.89$\pm$2.58 & 78.80$\pm$1.93 & 69.87$\pm$1.31 & 82.67$\pm$0.95 & 73.53$\pm$3.80 & 82.98$\pm$2.15 \\
\hline
Top-$B$ EPIG & 88.31$\pm$0.33 & \textbf{93.26$\pm$0.28} & 64.97$\pm$3.08 & 81.89$\pm$0.28 & 41.28$\pm$1.31 & \textbf{86.37$\pm$1.20} & 84.33$\pm$1.29 & 77.31$\pm$0.84 & 87.17$\pm$0.97 & \textbf{75.71$\pm$6.66} & 92.03$\pm$0.95 \\
PowerEPIG & 86.31$\pm$0.69 & 92.36$\pm$0.40 & 64.63$\pm$0.68 & 81.66$\pm$0.25 & \underline{\textbf{43.70$\pm$0.92}} & 81.03$\pm$1.47 & 81.53$\pm$1.39 & 75.76$\pm$1.01 & 82.70$\pm$0.80 & \textbf{78.93$\pm$5.00} & 84.60$\pm$1.62 \\
\hline
\end{tabular}
}
\caption{Bayesian AL comparison: $20+10\times20$. The reported number is final test accuracy averaged over 10 seeds with $\pm$ defining the $95\%$ confidence interval using Student's $t$-distribution. The highest mean accuracy is underlined, and all methods with means that are not significantly different ($p$-value greater than or equal to $0.05$) according to Welch's $t$-test are bolded.}
\label{tab:bs20_bayesian_ttest}
\end{table*}

\begin{table*}
\centering
\resizebox{\columnwidth}{!}{%
\begin{tabular}{l|ccccc|ccc|ccc}
\hline
Datasets & \multicolumn{5}{c|}{Tabular Datasets} & \multicolumn{3}{c|}{Text Datasets} & \multicolumn{3}{c}{Image Datasets} \\
Algorithm & Airline & Credit Card & Magic & Satellite & Vowels & Civil Comments & AG News & Yelp & CIFAR-10 & iWildCam & fMoW \\ \hline
Top-$B$ BALD & 86.94$\pm$0.75 & \textbf{93.28$\pm$0.20} & \textbf{66.97$\pm$1.74} & \textbf{81.44$\pm$0.56} & \textbf{41.80$\pm$0.87} & 79.63$\pm$1.71 & 77.04$\pm$1.65 & 69.61$\pm$1.46 & 79.35$\pm$1.12 & 74.41$\pm$3.02 & 81.79$\pm$1.46 \\
PowerBALD & 87.14$\pm$0.54 & 93.12$\pm$0.25 & \textbf{66.66$\pm$2.36} & \textbf{81.75$\pm$0.48} & \textbf{41.28$\pm$1.14} & 81.18$\pm$1.36 & 81.49$\pm$1.13 & 76.37$\pm$1.09 & 82.20$\pm$1.04 & \textbf{79.49$\pm$5.06} & 84.31$\pm$1.30 \\
SoftmaxBALD & 87.01$\pm$0.71 & 92.69$\pm$0.34 & 65.27$\pm$1.18 & \textbf{81.66$\pm$0.66} & \textbf{41.65$\pm$1.80} & 81.47$\pm$1.69 & 80.90$\pm$1.49 & 76.16$\pm$1.04 & 82.04$\pm$0.95 & \underline{\textbf{81.72$\pm$4.30}} & 84.18$\pm$1.37 \\
SoftRankBALD & 87.89$\pm$0.51 & \textbf{93.51$\pm$0.17} & 66.05$\pm$1.83 & \textbf{81.57$\pm$0.39} & \textbf{41.65$\pm$1.32} & 82.56$\pm$2.04 & 80.66$\pm$1.70 & 72.54$\pm$1.00 & 82.62$\pm$1.24 & \textbf{77.03$\pm$3.68} & 82.52$\pm$1.43 \\
BatchBALD & 86.69$\pm$0.72 & 93.27$\pm$0.09 & \textbf{67.98$\pm$1.89} & \textbf{81.39$\pm$0.77} & \textbf{41.80$\pm$0.91} & \textbf{82.63$\pm$2.68} & 78.66$\pm$1.43 & 70.82$\pm$0.87 & 81.90$\pm$1.39 & \textbf{76.92$\pm$3.27} & 84.52$\pm$1.45 \\
\hline
Top-$B$ EPIG & \textbf{88.20$\pm$0.37} & 93.15$\pm$0.27 & \textbf{66.68$\pm$1.90} & \textbf{82.01$\pm$0.43} & \underline{\textbf{42.51$\pm$1.30}} & \textbf{84.08$\pm$1.79} & \underline{\textbf{84.17$\pm$0.75}} & \textbf{77.12$\pm$1.11} & \textbf{84.73$\pm$0.71} & 74.72$\pm$5.64 & 90.37$\pm$1.28 \\
\hline
\end{tabular}
}
\caption{Bayesian AL comparison: $100+10\times10$. The reported number is final test accuracy averaged over 10 seeds with $\pm$ defining the $95\%$ confidence interval using Student's $t$-distribution. The highest mean accuracy is underlined, and all methods with means that are not significantly different ($p$-value greater than or equal to $0.05$) according to Welch's $t$-test are bolded.}
\label{tab:bs10_bayesian_ttest}
\end{table*}

\begin{table*}
\centering
\resizebox{\columnwidth}{!}{%
\begin{tabular}{l|ccccc|ccc|ccc}
\hline
Datasets & \multicolumn{5}{c|}{Tabular Datasets} & \multicolumn{3}{c|}{Text Datasets} & \multicolumn{3}{c}{Image Datasets} \\
Algorithm & Airline & Credit Card & Magic & Satellite & Vowels & Civil Comments & AG News & Yelp & CIFAR-10 & iWildCam & fMoW \\ \hline
Random & 87.24$\pm$0.52 & 92.84$\pm$0.29 & \textbf{65.77$\pm$2.10} & 82.40$\pm$0.48 & \underline{\textbf{43.55$\pm$0.52}} & 84.80$\pm$1.25 & 83.92$\pm$0.75 & 77.61$\pm$0.96 & 88.39$\pm$0.48 & 82.30$\pm$3.90 & 90.02$\pm$1.49 \\
Confidence & \textbf{89.35$\pm$0.33} & 92.71$\pm$0.41 & \textbf{66.71$\pm$1.74} & \textbf{82.44$\pm$0.59} & \textbf{43.33$\pm$0.36} & \underline{\textbf{89.67$\pm$0.74}} & \textbf{87.87$\pm$0.78} & \textbf{80.11$\pm$1.01} & \underline{\textbf{96.46$\pm$0.44}} & \underline{\textbf{90.24$\pm$3.00}} & \underline{\textbf{98.96$\pm$0.61}} \\
GLISTER & 87.43$\pm$0.73 & \textbf{93.42$\pm$0.35} & 62.25$\pm$4.61 & 80.86$\pm$0.65 & \textbf{43.16$\pm$0.62} & 84.13$\pm$1.53 & 82.12$\pm$0.89 & 73.07$\pm$1.02 & 91.65$\pm$1.01 & 79.29$\pm$2.99 & 91.69$\pm$3.28 \\
CoreSet & 85.87$\pm$0.91 & 93.05$\pm$0.24 & \textbf{66.28$\pm$1.53} & 81.02$\pm$0.43 & \textbf{43.46$\pm$0.32} & 71.58$\pm$1.42 & 74.59$\pm$1.46 & 69.26$\pm$1.12 & 73.75$\pm$0.71 & 71.95$\pm$3.12 & 73.35$\pm$3.33 \\
BADGE & 87.88$\pm$0.72 & 92.77$\pm$0.30 & \textbf{66.31$\pm$1.58} & 82.35$\pm$0.55 & \textbf{43.40$\pm$0.24} & 85.04$\pm$1.64 & 83.84$\pm$0.89 & 77.94$\pm$1.23 & 88.19$\pm$0.73 & 85.95$\pm$2.69 & 92.54$\pm$1.46 \\
GALAXY & \textbf{88.91$\pm$0.35} & 92.49$\pm$0.26 & \textbf{66.24$\pm$1.87} & \underline{\textbf{83.15$\pm$0.51}} & \textbf{43.38$\pm$0.35} & \textbf{89.55$\pm$1.10} & \textbf{87.24$\pm$0.65} & \textbf{80.28$\pm$0.79} & 94.80$\pm$0.61 & \textbf{86.34$\pm$3.13} & 94.50$\pm$3.22 \\
\hline
\end{tabular}
}
\caption{Heuristics AL comparison: $100+10\times20$. The reported number is final test accuracy averaged over 10 seeds with $\pm$ defining the $95\%$ confidence interval using Student's $t$-distribution. The highest mean accuracy is underlined, and all methods with means that are not significantly different ($p$-value greater than or equal to $0.05$) according to Welch's $t$-test are bolded.}
\label{tab:bs100_heuristic_ttest}
\end{table*}

\begin{table*}
\centering
\resizebox{\columnwidth}{!}{%
\begin{tabular}{l|ccccc|ccc|ccc}
\hline
Datasets & \multicolumn{5}{c|}{Tabular Datasets} & \multicolumn{3}{c|}{Text Datasets} & \multicolumn{3}{c}{Image Datasets} \\
Algorithm & Airline & Credit Card & Magic & Satellite & Vowels & Civil Comments & AG News & Yelp & CIFAR-10 & iWildCam & fMoW \\ \hline
Random & 86.17$\pm$0.65 & 92.41$\pm$0.38 & 65.92$\pm$1.35 & 80.99$\pm$0.72 & 42.01$\pm$1.34 & 82.63$\pm$1.60 & 80.69$\pm$0.89 & 76.47$\pm$0.83 & 83.47$\pm$1.16 & 78.58$\pm$2.95 & 85.47$\pm$1.33 \\
Confidence & \textbf{88.61$\pm$0.52} & 92.22$\pm$0.49 & 66.02$\pm$2.90 & 81.50$\pm$0.60 & \textbf{43.42$\pm$1.40} & 87.28$\pm$1.55 & \textbf{85.34$\pm$0.84} & \textbf{78.69$\pm$1.09} & \underline{\textbf{90.29$\pm$1.35}} & \underline{\textbf{86.65$\pm$2.30}} & \textbf{94.69$\pm$1.61} \\
GLISTER & 86.02$\pm$1.11 & 92.40$\pm$0.89 & \textbf{65.97$\pm$4.94} & 76.05$\pm$2.55 & 40.54$\pm$1.94 & 72.36$\pm$2.50 & 72.12$\pm$2.32 & 65.88$\pm$1.50 & 78.71$\pm$2.03 & 67.11$\pm$3.71 & 75.89$\pm$4.56 \\
CoreSet & 80.57$\pm$2.04 & \textbf{93.28$\pm$0.17} & \textbf{68.86$\pm$3.71} & 80.02$\pm$1.14 & 38.98$\pm$0.71 & 57.48$\pm$0.76 & 63.18$\pm$1.11 & 58.95$\pm$1.53 & 62.60$\pm$0.56 & 60.13$\pm$1.95 & 57.75$\pm$2.46 \\
BADGE & 87.09$\pm$0.58 & 93.18$\pm$0.23 & 64.85$\pm$2.31 & 81.42$\pm$0.52 & 41.84$\pm$1.54 & 83.07$\pm$1.42 & 81.27$\pm$0.85 & 75.50$\pm$0.82 & 83.59$\pm$0.77 & \textbf{83.97$\pm$2.22} & 85.81$\pm$1.48 \\
GALAXY & 88.33$\pm$0.50 & 92.44$\pm$0.45 & \textbf{66.84$\pm$1.55} & \textbf{82.43$\pm$0.47} & \textbf{43.25$\pm$0.96} & \underline{\textbf{89.32$\pm$0.86}} & 84.00$\pm$1.00 & \textbf{78.51$\pm$1.05} & \textbf{89.86$\pm$0.75} & 82.15$\pm$2.13 & 89.28$\pm$4.06 \\
\hline
\end{tabular}
}
\caption{Heuristics AL comparison: $20+10\times20$. The reported number is final test accuracy averaged over 10 seeds with $\pm$ defining the $95\%$ confidence interval using Student's $t$-distribution. The highest mean accuracy is underlined, and all methods with means that are not significantly different ($p$-value greater than or equal to $0.05$) according to Welch's $t$-test are bolded.}
\label{tab:bs20_heuristic_ttest}
\end{table*}

\begin{table*}
\centering
\resizebox{\columnwidth}{!}{%
\begin{tabular}{l|ccccc|ccc|ccc}
\hline
Datasets & \multicolumn{5}{c|}{Tabular Datasets} & \multicolumn{3}{c|}{Text Datasets} & \multicolumn{3}{c}{Image Datasets} \\
Algorithm & Airline & Credit Card & Magic & Satellite & Vowels & Civil Comments & AG News & Yelp & CIFAR-10 & iWildCam & fMoW \\ \hline
Random & 86.65$\pm$0.81 & 92.44$\pm$0.27 & \textbf{66.16$\pm$1.99} & 81.95$\pm$0.41 & \textbf{40.71$\pm$2.00} & 80.88$\pm$1.46 & 80.64$\pm$1.41 & 75.39$\pm$1.12 & 82.78$\pm$1.19 & \textbf{78.31$\pm$5.05} & 84.70$\pm$1.42 \\
Confidence & \textbf{88.15$\pm$0.66} & 92.82$\pm$0.41 & \textbf{66.48$\pm$1.99} & 81.14$\pm$0.66 & \textbf{42.47$\pm$1.56} & \textbf{86.43$\pm$1.79} & \textbf{83.78$\pm$1.09} & \textbf{77.31$\pm$1.10} & \textbf{87.31$\pm$0.87} & \underline{\textbf{83.55$\pm$3.11}} & 91.78$\pm$1.34 \\
GLISTER & 87.14$\pm$0.64 & \textbf{93.30$\pm$0.39} & 63.07$\pm$4.32 & 80.88$\pm$0.68 & 39.55$\pm$1.64 & 80.19$\pm$2.57 & 77.92$\pm$1.93 & 71.04$\pm$1.19 & 82.78$\pm$1.22 & 73.29$\pm$4.52 & 81.78$\pm$2.11 \\
CoreSet & 86.18$\pm$0.63 & \textbf{93.32$\pm$0.30} & 64.94$\pm$2.37 & 81.16$\pm$0.86 & 40.71$\pm$0.75 & 72.33$\pm$1.45 & 73.23$\pm$1.58 & 68.88$\pm$1.44 & 72.26$\pm$0.78 & 69.24$\pm$2.68 & 72.79$\pm$2.19 \\
BADGE & 86.94$\pm$0.46 & \textbf{92.88$\pm$0.48} & 65.70$\pm$2.09 & \textbf{82.26$\pm$0.51} & \textbf{41.80$\pm$2.02} & 79.76$\pm$1.79 & 79.44$\pm$1.33 & 75.19$\pm$1.08 & 81.44$\pm$0.75 & \textbf{80.40$\pm$3.32} & 82.56$\pm$1.49 \\
GALAXY & \textbf{87.82$\pm$0.73} & 92.15$\pm$0.28 & \textbf{67.30$\pm$2.66} & \underline{\textbf{82.91$\pm$0.54}} & \textbf{41.30$\pm$1.41} & \underline{\textbf{87.77$\pm$1.02}} & \textbf{83.59$\pm$1.34} & \underline{\textbf{78.09$\pm$1.23}} & \underline{\textbf{87.90$\pm$0.98}} & \textbf{82.95$\pm$4.48} & 88.44$\pm$4.32 \\
\hline
\end{tabular}
}
\caption{Heuristics AL comparison: $100+10\times10$. The reported number is final test accuracy averaged over 10 seeds with $\pm$ defining the $95\%$ confidence interval using Student's $t$-distribution. The highest mean accuracy is underlined, and all methods with means that are not significantly different ($p$-value greater than or equal to $0.05$) according to Welch's $t$-test are bolded.}
\label{tab:bs10_heuristic_ttest}
\end{table*}

\begin{table*}
\centering
\resizebox{\columnwidth}{!}{%
\begin{tabular}{l|cccccccccc}
\hline
Datasets & \multicolumn{10}{c}{One-vs-all CIFAR-10} \\
Algorithm & 0 & 1 & 2 & 3 & 4 & 5 & 6 & 7 & 8 & 9 \\ \hline
Top-$B$ BALD & 77.93$\pm$1.06 & 82.13$\pm$0.89 & 76.68$\pm$1.52 & 77.13$\pm$1.02 & 82.07$\pm$1.14 & 79.52$\pm$0.91 & 79.18$\pm$0.92 & 83.30$\pm$0.79 & 80.92$\pm$1.11 & 80.79$\pm$0.85 \\
PowerBALD & 83.22$\pm$0.72 & 84.28$\pm$0.48 & 84.06$\pm$0.94 & 81.79$\pm$0.74 & 82.14$\pm$1.45 & 82.03$\pm$0.69 & 84.20$\pm$1.07 & 84.97$\pm$0.73 & 84.42$\pm$0.64 & 84.32$\pm$0.60 \\
SoftmaxBALD & 83.86$\pm$1.11 & 83.43$\pm$0.54 & 83.34$\pm$0.72 & 81.38$\pm$1.26 & 81.64$\pm$1.16 & 82.76$\pm$0.76 & 84.00$\pm$0.95 & 84.07$\pm$1.11 & 84.04$\pm$0.88 & 83.87$\pm$0.84 \\
SoftRankBALD & 82.67$\pm$0.95 & 85.70$\pm$0.94 & 81.82$\pm$0.91 & 80.57$\pm$0.71 & 84.24$\pm$1.02 & 82.67$\pm$0.89 & 83.24$\pm$0.95 & 86.18$\pm$1.19 & 84.28$\pm$1.10 & 84.96$\pm$0.73 \\
\hline
Top-$B$ EPIG & 87.17$\pm$0.97 & 87.41$\pm$0.84 & 86.06$\pm$1.22 & 85.17$\pm$0.91 & 85.42$\pm$0.73 & 86.60$\pm$0.69 & 88.00$\pm$0.75 & 88.65$\pm$1.09 & 88.25$\pm$1.04 & 87.60$\pm$0.83 \\
PowerEPIG & 82.70$\pm$0.80 & 83.12$\pm$0.89 & 81.83$\pm$0.82 & 81.15$\pm$0.96 & 82.27$\pm$0.90 & 82.43$\pm$0.75 & 83.31$\pm$0.80 & 83.11$\pm$0.87 & 83.99$\pm$1.10 & 82.86$\pm$1.21 \\
\hline
\end{tabular}
}
\caption{Final test accuracy of Bayesian AL algorithms with Bayesian Logistic Regression on One-vs-all CIFAR-10, where each of the 10 iterations has a labeling budget of 20 samples, starting with random initialization of 20 samples.}
\label{tab:cifar10_ova_bs20_bayesian_ttest}
\end{table*}

\begin{table*}
\centering
\resizebox{\columnwidth}{!}{%
\begin{tabular}{l|cccccccccc}
\hline
Datasets & \multicolumn{10}{c}{One-vs-all CIFAR-10} \\
Algorithm & 0 & 1 & 2 & 3 & 4 & 5 & 6 & 7 & 8 & 9 \\ \hline
Top-$B$ BALD & 88.25$\pm$0.93 & 92.87$\pm$0.64 & 86.98$\pm$0.95 & 85.79$\pm$1.10 & 91.68$\pm$1.04 & 89.07$\pm$1.04 & 89.23$\pm$0.59 & 94.05$\pm$1.17 & 91.70$\pm$0.65 & 91.84$\pm$0.83 \\
PowerBALD & 90.06$\pm$0.75 & 90.22$\pm$0.89 & 89.17$\pm$1.39 & 87.22$\pm$1.09 & 88.71$\pm$0.87 & 87.95$\pm$1.36 & 89.53$\pm$1.15 & 90.54$\pm$0.87 & 90.95$\pm$1.00 & 89.13$\pm$0.70 \\
SoftmaxBALD & 88.90$\pm$0.79 & 89.88$\pm$1.07 & 88.78$\pm$0.96 & 87.46$\pm$0.71 & 87.58$\pm$1.30 & 88.07$\pm$1.14 & 89.12$\pm$0.54 & 89.72$\pm$1.21 & 90.37$\pm$0.78 & 89.96$\pm$1.07 \\
SoftRankBALD & 91.74$\pm$0.86 & 94.65$\pm$0.49 & 91.00$\pm$0.92 & 88.83$\pm$0.64 & 91.84$\pm$1.14 & 91.46$\pm$0.99 & 91.65$\pm$0.71 & 94.99$\pm$0.55 & 93.45$\pm$0.98 & 93.11$\pm$0.63 \\
\hline
Top-$B$ EPIG & 93.82$\pm$1.02 & 94.50$\pm$0.95 & 93.23$\pm$1.09 & 91.24$\pm$0.71 & 92.12$\pm$0.85 & 92.64$\pm$0.62 & 94.84$\pm$0.41 & 94.48$\pm$0.88 & 94.98$\pm$0.52 & 94.74$\pm$0.87 \\
PowerEPIG & 88.54$\pm$0.73 & 89.16$\pm$0.59 & 87.73$\pm$0.85 & 86.30$\pm$0.84 & 86.88$\pm$0.76 & 87.28$\pm$0.83 & 88.37$\pm$1.02 & 88.36$\pm$0.96 & 89.17$\pm$0.46 & 89.15$\pm$0.60 \\
\hline
\end{tabular}
}
\caption{Final test accuracy of Bayesian AL algorithms with Bayesian Logistic Regression on One-vs-all CIFAR-10, where each of the 10 iterations has a labeling budget of 20 samples, starting with random initialization of 100 samples.}
\label{tab:cifar10_ova_bs100_bayesian_ttest}
\end{table*}

\begin{table*}
\centering
\resizebox{\columnwidth}{!}{%
\begin{tabular}{l|cccccccccc}
\hline
Datasets & \multicolumn{10}{c}{One-vs-all CIFAR-10} \\
Algorithm & 0 & 1 & 2 & 3 & 4 & 5 & 6 & 7 & 8 & 9 \\ \hline
Random & 83.47$\pm$1.16 & 83.93$\pm$0.75 & 82.99$\pm$1.16 & 81.90$\pm$0.75 & 82.58$\pm$1.05 & 82.60$\pm$1.21 & 83.94$\pm$0.97 & 84.36$\pm$0.89 & 83.70$\pm$0.90 & 83.51$\pm$1.03 \\
Confidence & \underline{\textbf{90.29$\pm$1.35}} & \underline{\textbf{91.16$\pm$0.97}} & \textbf{89.48$\pm$0.70} & \textbf{87.04$\pm$1.07} & \textbf{88.47$\pm$0.67} & \textbf{88.55$\pm$0.96} & \textbf{90.85$\pm$0.73} & \textbf{91.32$\pm$0.93} & \textbf{91.66$\pm$0.55} & \textbf{90.63$\pm$0.82} \\
GLISTER & 78.71$\pm$2.03 & 81.66$\pm$1.70 & 77.62$\pm$1.38 & 78.59$\pm$0.85 & 79.34$\pm$1.03 & 79.14$\pm$2.04 & 80.82$\pm$2.01 & 80.57$\pm$0.91 & 81.00$\pm$1.89 & 79.81$\pm$2.81 \\
CoreSet & 62.60$\pm$0.56 & 63.46$\pm$0.62 & 61.54$\pm$0.76 & 62.10$\pm$0.53 & 63.35$\pm$0.64 & 63.05$\pm$0.46 & 64.33$\pm$0.66 & 62.83$\pm$0.32 & 63.24$\pm$0.71 & 62.94$\pm$0.59 \\
BADGE & 83.59$\pm$0.77 & 83.15$\pm$0.99 & 82.45$\pm$1.21 & 81.40$\pm$0.94 & 82.07$\pm$0.92 & 82.39$\pm$1.08 & 83.25$\pm$1.12 & 84.31$\pm$0.94 & 84.25$\pm$1.02 & 84.02$\pm$1.28 \\
GALAXY & \textbf{89.86$\pm$0.75} & \textbf{90.66$\pm$1.16} & \underline{\textbf{89.77$\pm$0.87}} & \textbf{87.76$\pm$0.97} & \underline{\textbf{89.21$\pm$0.49}} & \underline{\textbf{89.65$\pm$1.09}} & \underline{\textbf{91.91$\pm$0.97}} & \underline{\textbf{91.90$\pm$1.23}} & \underline{\textbf{92.15$\pm$0.99}} & \underline{\textbf{91.70$\pm$1.09}} \\
\hline
ParBaLS-MAP EPIG & 88.67$\pm$0.76 & 89.48$\pm$0.75 & 87.48$\pm$0.66 & 85.55$\pm$0.75 & 86.98$\pm$0.71 & 86.91$\pm$0.72 & 88.49$\pm$0.80 & 89.36$\pm$0.84 & 89.36$\pm$0.71 & 88.76$\pm$0.77 \\
ParBaLS EPIG & \textbf{88.95$\pm$0.70} & 89.64$\pm$0.95 & 88.10$\pm$1.08 & \underline{\textbf{88.01$\pm$0.54}} & 87.96$\pm$0.72 & 88.23$\pm$0.68 & 89.83$\pm$0.73 & 90.53$\pm$0.70 & 89.95$\pm$0.88 & 89.49$\pm$0.70 \\
\hline
\end{tabular}
}
\caption{Final test accuracy of ParBaLS and heuristic baselines with Bayesian Logistic Regression on One-vs-all CIFAR-10, where each of the 10 iterations has a labeling budget of 20 samples, starting with random initialization of 20 samples.}
\label{tab:cifar10_ova_bs20_heuristic_ttest}
\end{table*}

\begin{table*}
\centering
\resizebox{\columnwidth}{!}{%
\begin{tabular}{l|cccccccccc}
\hline
Datasets & \multicolumn{10}{c}{One-vs-all CIFAR-10} \\
Algorithm & 0 & 1 & 2 & 3 & 4 & 5 & 6 & 7 & 8 & 9 \\ \hline
Random & 88.39$\pm$0.48 & 89.21$\pm$0.91 & 87.62$\pm$0.84 & 86.23$\pm$0.55 & 87.60$\pm$0.87 & 86.86$\pm$0.91 & 88.78$\pm$1.09 & 88.69$\pm$0.87 & 89.14$\pm$0.77 & 88.60$\pm$1.06 \\
Confidence & \underline{\textbf{96.46$\pm$0.44}} & \underline{\textbf{97.55$\pm$0.43}} & \underline{\textbf{95.44$\pm$0.64}} & \underline{\textbf{93.21$\pm$0.74}} & \underline{\textbf{94.75$\pm$0.53}} & \underline{\textbf{94.69$\pm$0.60}} & \underline{\textbf{96.97$\pm$0.22}} & \underline{\textbf{97.82$\pm$0.44}} & \underline{\textbf{97.87$\pm$0.40}} & \underline{\textbf{96.82$\pm$0.33}} \\
GLISTER & 91.65$\pm$1.01 & 92.88$\pm$0.73 & 89.61$\pm$1.92 & 89.22$\pm$1.21 & 91.22$\pm$1.14 & 89.46$\pm$1.33 & 92.36$\pm$1.12 & 93.78$\pm$1.26 & 92.01$\pm$1.73 & 92.75$\pm$0.99 \\
CoreSet & 73.75$\pm$0.71 & 75.39$\pm$0.90 & 73.36$\pm$0.74 & 72.20$\pm$0.95 & 74.11$\pm$0.67 & 74.42$\pm$0.93 & 74.87$\pm$0.89 & 74.44$\pm$0.90 & 74.69$\pm$0.67 & 73.87$\pm$0.87 \\
BADGE & 88.19$\pm$0.73 & 89.96$\pm$0.98 & 87.66$\pm$0.99 & 86.47$\pm$0.93 & 88.06$\pm$0.99 & 88.12$\pm$0.61 & 88.87$\pm$0.67 & 90.70$\pm$0.89 & 89.95$\pm$1.00 & 89.09$\pm$0.44 \\
GALAXY & 94.80$\pm$0.61 & 95.71$\pm$0.60 & 94.05$\pm$0.49 & 91.98$\pm$0.68 & 93.24$\pm$0.69 & 93.41$\pm$0.95 & 95.68$\pm$0.58 & 96.27$\pm$0.65 & 96.67$\pm$0.39 & 95.12$\pm$0.75 \\
\hline
ParBaLS-MAP EPIG & 94.93$\pm$0.68 & 95.75$\pm$0.44 & 93.46$\pm$0.70 & \textbf{92.44$\pm$0.47} & 93.12$\pm$0.57 & 93.17$\pm$0.65 & 95.12$\pm$0.55 & 95.81$\pm$0.70 & 96.07$\pm$0.44 & 95.28$\pm$0.55 \\
ParBaLS EPIG & 95.40$\pm$0.80 & 96.69$\pm$0.28 & \textbf{94.84$\pm$0.89} & \textbf{93.08$\pm$0.63} & 94.01$\pm$0.59 & \textbf{94.39$\pm$0.37} & 96.06$\pm$0.41 & 96.69$\pm$0.51 & 96.92$\pm$0.44 & \textbf{96.60$\pm$0.23} \\
\hline
\end{tabular}
}
\caption{Final test accuracy of ParBaLS and heuristic baselines with Bayesian Logistic Regression on One-vs-all CIFAR-10, where each of the 10 iterations has a labeling budget of 20 samples, starting with random initialization of 100 samples.}
\label{tab:cifar10_ova_bs100_heuristic_ttest}
\end{table*}

\begin{table*}
\centering
\resizebox{\columnwidth}{!}{%
\begin{tabular}{l|ccccc}
\hline
Datasets & \multicolumn{5}{c}{One-vs-all iWildCam} \\
Algorithm & 0 & 1 & 2 & 3 & 4 \\ \hline
Top-$B$ BALD & 66.73$\pm$4.06 & 74.30$\pm$2.91 & 76.66$\pm$3.20 & 78.34$\pm$2.10 & 75.41$\pm$3.17 \\
PowerBALD & \textbf{79.84$\pm$4.44} & \textbf{83.63$\pm$4.00} & \textbf{82.14$\pm$3.03} & \textbf{86.92$\pm$4.52} & \textbf{82.66$\pm$3.74} \\
SoftmaxBALD & \textbf{81.31$\pm$4.93} & \textbf{82.49$\pm$3.18} & \textbf{82.78$\pm$2.12} & \textbf{85.52$\pm$4.76} & \textbf{82.28$\pm$3.63} \\
SoftRankBALD & 73.53$\pm$3.80 & \textbf{83.37$\pm$3.51} & \underline{\textbf{83.00$\pm$4.28}} & 84.90$\pm$3.68 & \textbf{83.49$\pm$2.48} \\
\hline
Top-$B$ EPIG & \textbf{75.71$\pm$6.66} & 78.56$\pm$5.40 & 74.86$\pm$4.80 & 83.96$\pm$4.15 & \textbf{80.81$\pm$4.45} \\
PowerEPIG & \textbf{78.93$\pm$5.00} & 80.25$\pm$4.36 & \textbf{77.83$\pm$5.35} & 80.71$\pm$4.50 & 78.59$\pm$4.46 \\
SoftmaxEPIG & 76.94$\pm$4.04 & \underline{\textbf{86.22$\pm$3.77}} & \textbf{82.39$\pm$4.61} & \textbf{88.77$\pm$2.54} & \textbf{83.52$\pm$3.66} \\
SoftRankEPIG & \textbf{78.01$\pm$5.30} & \textbf{82.05$\pm$4.35} & \textbf{79.05$\pm$4.82} & 84.65$\pm$4.65 & \underline{\textbf{84.90$\pm$3.70}} \\
\hline
ParBaLS-MAP EPIG & \textbf{80.26$\pm$3.30} & \textbf{85.13$\pm$2.36} & \textbf{78.86$\pm$4.63} & \underline{\textbf{90.27$\pm$2.02}} & \textbf{81.17$\pm$3.60} \\
ParBaLS EPIG & \underline{\textbf{82.98$\pm$4.20}} & \textbf{83.15$\pm$3.57} & \textbf{78.71$\pm$2.43} & \textbf{87.84$\pm$3.67} & \textbf{83.45$\pm$3.31} \\
\hline
\end{tabular}
}
\caption{Final test accuracy of Bayesian AL algorithms with Bayesian Logistic Regression on One-vs-all iWildCam, where each of the 10 iterations has a labeling budget of 20 samples, starting with random initialization of 20 samples.}
\label{tab:iwildcam_ood_bs20_bayesian_ttest}
\end{table*}

\begin{table*}
\centering
\resizebox{\columnwidth}{!}{%
\begin{tabular}{l|ccccc}
\hline
Datasets & \multicolumn{5}{c}{One-vs-all iWildCam} \\
Algorithm & 0 & 1 & 2 & 3 & 4 \\ \hline
Top-$B$ BALD & 78.62$\pm$2.83 & 87.80$\pm$2.69 & \textbf{89.33$\pm$3.67} & \textbf{93.68$\pm$2.35} & \textbf{90.42$\pm$1.99} \\
PowerBALD & 82.93$\pm$4.62 & 86.18$\pm$3.40 & 83.73$\pm$5.29 & 91.36$\pm$2.64 & \textbf{87.60$\pm$2.58} \\
SoftmaxBALD & 81.99$\pm$4.63 & 85.45$\pm$3.94 & 84.44$\pm$5.00 & 91.02$\pm$3.04 & \textbf{87.43$\pm$3.46} \\
SoftRankBALD & 85.27$\pm$1.23 & \textbf{89.87$\pm$2.99} & \underline{\textbf{90.97$\pm$3.45}} & \textbf{94.98$\pm$4.25} & \underline{\textbf{91.25$\pm$3.03}} \\
\hline
Top-$B$ EPIG & 85.80$\pm$3.00 & 87.68$\pm$2.75 & 81.24$\pm$4.56 & 91.66$\pm$3.95 & \textbf{86.56$\pm$4.49} \\
PowerEPIG & 84.12$\pm$3.38 & 83.73$\pm$2.77 & 82.50$\pm$2.89 & 88.36$\pm$2.93 & 84.03$\pm$3.75 \\
SoftmaxEPIG & 86.15$\pm$1.56 & \textbf{88.41$\pm$3.71} & 83.30$\pm$5.20 & 91.08$\pm$4.07 & \textbf{88.27$\pm$3.09} \\
SoftRankEPIG & 86.50$\pm$2.39 & 86.62$\pm$4.76 & \textbf{85.25$\pm$5.83} & \textbf{93.55$\pm$2.03} & \textbf{88.90$\pm$2.97} \\
\hline
ParBaLS-MAP EPIG & 86.17$\pm$2.59 & \textbf{89.02$\pm$3.21} & \textbf{87.02$\pm$4.07} & \underline{\textbf{96.11$\pm$2.04}} & \textbf{91.18$\pm$2.70} \\
ParBaLS EPIG & \underline{\textbf{89.77$\pm$1.46}} & \underline{\textbf{91.84$\pm$2.02}} & 86.72$\pm$2.87 & \textbf{95.68$\pm$1.26} & \textbf{91.12$\pm$1.78} \\
\hline
\end{tabular}
}
\caption{Final test accuracy of Bayesian AL algorithms with Bayesian Logistic Regression on One-vs-all iWildCam, where each of the 10 iterations has a labeling budget of 20 samples, starting with random initialization of 100 samples.}
\label{tab:iwildcam_ood_bs100_bayesian_ttest}
\end{table*}

\begin{table*}
\centering
\resizebox{\columnwidth}{!}{%
\begin{tabular}{l|ccccc}
\hline
Datasets & \multicolumn{5}{c}{One-vs-all iWildCam} \\
Algorithm & 0 & 1 & 2 & 3 & 4 \\ \hline
Random & 78.58$\pm$2.95 & 81.47$\pm$4.14 & 80.23$\pm$5.51 & 85.29$\pm$4.27 & 83.42$\pm$3.90 \\
Confidence & \underline{\textbf{86.65$\pm$2.30}} & \textbf{92.47$\pm$1.76} & \textbf{90.27$\pm$2.09} & \underline{\textbf{95.02$\pm$3.61}} & \textbf{92.41$\pm$2.02} \\
GLISTER & 67.11$\pm$3.71 & 70.73$\pm$4.06 & 72.56$\pm$5.24 & 74.70$\pm$3.83 & 74.15$\pm$3.04 \\
CoreSet & 60.13$\pm$1.95 & 63.37$\pm$3.28 & 57.61$\pm$5.08 & 60.50$\pm$4.09 & 53.88$\pm$3.31 \\
BADGE & \textbf{83.97$\pm$2.22} & 83.01$\pm$2.69 & 80.29$\pm$3.62 & 87.80$\pm$3.63 & 86.10$\pm$3.91 \\
GALAXY & 82.15$\pm$2.13 & \underline{\textbf{94.14$\pm$1.24}} & \underline{\textbf{92.94$\pm$2.30}} & 84.98$\pm$4.31 & \underline{\textbf{94.41$\pm$2.68}} \\
\hline
ParBaLS-MAP EPIG & 80.26$\pm$3.30 & 85.13$\pm$2.36 & 78.86$\pm$4.63 & 90.27$\pm$2.02 & 81.17$\pm$3.60 \\
ParBaLS EPIG & \textbf{82.98$\pm$4.20} & 83.15$\pm$3.57 & 78.71$\pm$2.43 & 87.84$\pm$3.67 & 83.45$\pm$3.31 \\
\hline
\end{tabular}
}
\caption{Final test accuracy of ParBaLS and heuristic baselines with Bayesian Logistic Regression on One-vs-all iWildCam, where each of the 10 iterations has a labeling budget of 20 samples, starting with random initialization of 20 samples.}
\label{tab:iwildcam_ood_bs20_heuristic_ttest}
\end{table*}

\begin{table*}
\centering
\resizebox{\columnwidth}{!}{%
\begin{tabular}{l|ccccc}
\hline
Datasets & \multicolumn{5}{c}{One-vs-all iWildCam} \\
Algorithm & 0 & 1 & 2 & 3 & 4 \\ \hline
Random & 82.30$\pm$3.90 & 83.00$\pm$4.90 & 81.70$\pm$5.83 & 88.41$\pm$4.18 & 85.87$\pm$2.98 \\
Confidence & \underline{\textbf{90.24$\pm$3.00}} & \underline{\textbf{96.69$\pm$2.03}} & \underline{\textbf{94.12$\pm$1.71}} & \underline{\textbf{99.78$\pm$0.02}} & \textbf{96.51$\pm$1.68} \\
GLISTER & 79.29$\pm$2.99 & 85.97$\pm$2.72 & 88.90$\pm$3.35 & 92.28$\pm$2.48 & 89.59$\pm$2.13 \\
CoreSet & 71.95$\pm$3.12 & 73.24$\pm$3.24 & 71.84$\pm$5.41 & 73.57$\pm$4.55 & 66.75$\pm$4.65 \\
BADGE & 85.95$\pm$2.69 & 87.23$\pm$4.01 & 86.00$\pm$5.70 & 92.35$\pm$3.98 & 88.90$\pm$3.14 \\
GALAXY & \textbf{86.34$\pm$3.13} & \textbf{93.47$\pm$4.39} & \textbf{93.74$\pm$1.73} & 89.14$\pm$5.39 & \underline{\textbf{96.53$\pm$1.60}} \\
\hline
ParBaLS-MAP EPIG & 86.17$\pm$2.59 & 89.02$\pm$3.21 & 87.02$\pm$4.07 & 96.11$\pm$2.04 & 91.18$\pm$2.70 \\
ParBaLS EPIG & \textbf{89.77$\pm$1.46} & 91.84$\pm$2.02 & 86.72$\pm$2.87 & 95.68$\pm$1.26 & 91.12$\pm$1.78 \\
\hline
\end{tabular}
}
\caption{Final test accuracy of ParBaLS and heuristic baselines with Bayesian Logistic Regression on One-vs-all iWildCam, where each of the 10 iterations has a labeling budget of 20 samples, starting with random initialization of 100 samples.}
\label{tab:iwildcam_ood_bs100_heuristic_ttest}
\end{table*}

\begin{table*}
\centering
\resizebox{\columnwidth}{!}{%
\begin{tabular}{l|ccccc}
\hline
Datasets & \multicolumn{5}{c}{One-vs-all fMoW} \\
Algorithm & 0 & 1 & 2 & 3 & 4 \\ \hline
Top-$B$ BALD & 75.47$\pm$2.40 & 74.41$\pm$1.92 & 73.43$\pm$2.37 & 71.15$\pm$1.24 & 74.55$\pm$1.91 \\
PowerBALD & 86.60$\pm$2.14 & 84.77$\pm$1.35 & 85.61$\pm$1.80 & 83.91$\pm$1.64 & 86.96$\pm$1.52 \\
SoftmaxBALD & 87.06$\pm$1.78 & 84.85$\pm$1.37 & 85.04$\pm$1.38 & 83.85$\pm$1.40 & 86.21$\pm$1.54 \\
SoftRankBALD & 82.98$\pm$2.15 & 81.86$\pm$1.66 & 81.18$\pm$2.22 & 79.76$\pm$2.60 & 83.88$\pm$2.52 \\
\hline
Top-$B$ EPIG & 92.03$\pm$0.95 & 90.12$\pm$0.64 & 91.27$\pm$0.99 & 87.99$\pm$1.24 & 91.91$\pm$0.90 \\
PowerEPIG & 84.60$\pm$1.62 & 82.66$\pm$1.59 & 82.91$\pm$1.60 & 81.59$\pm$1.86 & 84.39$\pm$1.83 \\
SoftmaxEPIG & 87.27$\pm$1.02 & 86.80$\pm$1.56 & 86.59$\pm$1.05 & 84.61$\pm$1.47 & 87.70$\pm$1.02 \\
SoftRankEPIG & 91.20$\pm$0.88 & 89.45$\pm$1.12 & 89.99$\pm$0.96 & 87.77$\pm$1.38 & 91.42$\pm$0.90 \\
\hline
ParBaLS-MAP EPIG & 93.51$\pm$0.78 & 91.04$\pm$1.10 & 92.88$\pm$0.77 & \textbf{89.96$\pm$1.57} & 93.42$\pm$0.81 \\
ParBaLS EPIG & \underline{\textbf{95.45$\pm$0.76}} & \underline{\textbf{93.53$\pm$0.61}} & \underline{\textbf{94.22$\pm$0.80}} & \underline{\textbf{91.52$\pm$1.15}} & \underline{\textbf{94.96$\pm$0.53}} \\
\hline
\end{tabular}
}
\caption{Final test accuracy of Bayesian AL algorithms with Bayesian Logistic Regression on One-vs-all fMoW, where each of the 10 iterations has a labeling budget of 20 samples, starting with random initialization of 20 samples.}
\label{tab:fmow_ood_bs20_bayesian_ttest}
\end{table*}

\begin{table*}
\centering
\resizebox{\columnwidth}{!}{%
\begin{tabular}{l|ccccc}
\hline
Datasets & \multicolumn{5}{c}{One-vs-all fMoW} \\
Algorithm & 0 & 1 & 2 & 3 & 4 \\ \hline
Top-$B$ BALD & 91.78$\pm$3.84 & 88.81$\pm$2.49 & 89.80$\pm$2.10 & 84.56$\pm$2.98 & 91.57$\pm$2.80 \\
PowerBALD & 91.42$\pm$1.18 & 89.52$\pm$1.91 & 90.24$\pm$0.94 & 88.55$\pm$1.37 & 91.11$\pm$1.19 \\
SoftmaxBALD & 90.44$\pm$1.51 & 88.33$\pm$1.36 & 89.09$\pm$1.17 & 88.08$\pm$1.54 & 90.64$\pm$1.06 \\
SoftRankBALD & 94.12$\pm$2.02 & 90.59$\pm$2.79 & 92.58$\pm$1.52 & 88.80$\pm$2.16 & 94.44$\pm$2.08 \\
\hline
Top-$B$ EPIG & 97.06$\pm$0.72 & 94.67$\pm$0.68 & 96.25$\pm$0.28 & \textbf{93.80$\pm$0.88} & 96.96$\pm$0.59 \\
PowerEPIG & 90.59$\pm$0.80 & 88.54$\pm$1.05 & 89.75$\pm$0.74 & 86.52$\pm$1.23 & 90.60$\pm$0.85 \\
SoftmaxEPIG & 92.66$\pm$0.68 & 90.37$\pm$0.93 & 92.21$\pm$0.94 & 88.46$\pm$1.39 & 92.12$\pm$1.18 \\
SoftRankEPIG & 96.09$\pm$0.71 & 93.94$\pm$0.85 & 94.97$\pm$0.97 & 91.96$\pm$1.46 & 95.92$\pm$0.57 \\
\hline
ParBaLS-MAP EPIG & \textbf{97.87$\pm$0.44} & \textbf{95.56$\pm$0.55} & \textbf{96.66$\pm$1.05} & \textbf{93.41$\pm$1.00} & 97.38$\pm$0.40 \\
ParBaLS EPIG & \underline{\textbf{98.41$\pm$0.49}} & \underline{\textbf{96.40$\pm$0.71}} & \underline{\textbf{97.65$\pm$0.52}} & \underline{\textbf{94.40$\pm$0.92}} & \underline{\textbf{98.42$\pm$0.47}} \\
\hline
\end{tabular}
}
\caption{Final test accuracy of Bayesian AL algorithms with Bayesian Logistic Regression on One-vs-all fMoW, where each of the 10 iterations has a labeling budget of 20 samples, starting with random initialization of 100 samples.}
\label{tab:fmow_ood_bs100_bayesian_ttest}
\end{table*}

\begin{table*}
\centering
\resizebox{\columnwidth}{!}{%
\begin{tabular}{l|ccccc}
\hline
Datasets & \multicolumn{5}{c}{One-vs-all fMoW} \\
Algorithm & 0 & 1 & 2 & 3 & 4 \\ \hline
Random & 85.47$\pm$1.33 & 83.47$\pm$1.38 & 84.42$\pm$1.09 & 82.61$\pm$1.91 & 85.39$\pm$1.33 \\
Confidence & \textbf{94.69$\pm$1.61} & 91.84$\pm$1.06 & 93.09$\pm$1.44 & 91.44$\pm$1.41 & \textbf{94.65$\pm$1.41} \\
GLISTER & 75.89$\pm$4.56 & 74.36$\pm$4.60 & 73.75$\pm$4.43 & 71.15$\pm$4.58 & 75.09$\pm$3.97 \\
CoreSet & 57.75$\pm$2.46 & 56.98$\pm$2.28 & 58.71$\pm$2.14 & 58.43$\pm$2.21 & 57.29$\pm$2.11 \\
BADGE & 85.81$\pm$1.48 & 83.89$\pm$1.11 & 85.21$\pm$1.21 & 82.97$\pm$1.82 & 86.48$\pm$0.94 \\
GALAXY & 89.28$\pm$4.06 & \underline{\textbf{95.64$\pm$2.55}} & \underline{\textbf{95.90$\pm$1.61}} & \underline{\textbf{95.60$\pm$1.15}} & \textbf{90.96$\pm$4.85} \\
\hline
ParBaLS-MAP EPIG & 93.51$\pm$0.78 & 91.04$\pm$1.10 & 92.88$\pm$0.77 & 89.96$\pm$1.57 & 93.42$\pm$0.81 \\
ParBaLS EPIG & \underline{\textbf{95.45$\pm$0.76}} & \textbf{93.53$\pm$0.61} & \textbf{94.22$\pm$0.80} & 91.52$\pm$1.15 & \underline{\textbf{94.96$\pm$0.53}} \\
\hline
\end{tabular}
}
\caption{Final test accuracy of ParBaLS and heuristic baselines with Bayesian Logistic Regression on One-vs-all fMoW, where each of the 10 iterations has a labeling budget of 20 samples, starting with random initialization of 20 samples.}
\label{tab:fmow_ood_bs20_heuristic_ttest}
\end{table*}

\begin{table*}
\centering
\resizebox{\columnwidth}{!}{%
\begin{tabular}{l|ccccc}
\hline
Datasets & \multicolumn{5}{c}{One-vs-all fMoW} \\
Algorithm & 0 & 1 & 2 & 3 & 4 \\ \hline
Random & 90.02$\pm$1.49 & 88.00$\pm$2.00 & 89.04$\pm$1.77 & 86.66$\pm$2.10 & 89.89$\pm$1.44 \\
Confidence & \underline{\textbf{98.96$\pm$0.61}} & \underline{\textbf{97.42$\pm$0.82}} & \underline{\textbf{98.20$\pm$0.43}} & \textbf{95.84$\pm$1.38} & \underline{\textbf{99.20$\pm$0.27}} \\
GLISTER & 91.69$\pm$3.28 & 89.51$\pm$3.01 & 89.75$\pm$3.36 & 87.01$\pm$3.99 & 92.18$\pm$3.25 \\
CoreSet & 73.35$\pm$3.33 & 72.14$\pm$3.12 & 73.52$\pm$2.69 & 71.78$\pm$3.13 & 72.91$\pm$2.95 \\
BADGE & 92.54$\pm$1.46 & 90.66$\pm$0.95 & 91.13$\pm$1.01 & 88.84$\pm$1.41 & 91.50$\pm$1.04 \\
GALAXY & 94.50$\pm$3.22 & \textbf{96.89$\pm$1.96} & \textbf{96.42$\pm$2.62} & \underline{\textbf{96.47$\pm$1.80}} & \textbf{97.00$\pm$2.35} \\
\hline
ParBaLS-MAP EPIG & 97.87$\pm$0.44 & 95.56$\pm$0.55 & 96.66$\pm$1.05 & 93.41$\pm$1.00 & 97.38$\pm$0.40 \\
ParBaLS EPIG & \textbf{98.41$\pm$0.49} & 96.40$\pm$0.71 & \textbf{97.65$\pm$0.52} & 94.40$\pm$0.92 & 98.42$\pm$0.47 \\
\hline
\end{tabular}
}
\caption{Final test accuracy of ParBaLS and heuristic baselines with Bayesian Logistic Regression on One-vs-all fMoW, where each of the 10 iterations has a labeling budget of 20 samples, starting with random initialization of 100 samples.}
\label{tab:fmow_ood_bs100_heuristic_ttest}
\end{table*}

\begin{table*}
\centering
\resizebox{\columnwidth}{!}{%
\begin{tabular}{l|cccc}
\hline
Datasets & \multicolumn{4}{c}{One-vs-all AG News} \\
Algorithm & 0 & 1 & 2 & 3 \\ \hline
Top-$B$ BALD & 69.93$\pm$2.01 & 76.70$\pm$1.37 & 69.21$\pm$1.30 & 68.83$\pm$1.10 \\
PowerBALD & 82.21$\pm$1.08 & 88.81$\pm$1.09 & 77.43$\pm$1.50 & 78.93$\pm$1.16 \\
SoftmaxBALD & 81.96$\pm$1.41 & 87.83$\pm$1.36 & 78.30$\pm$0.98 & 79.47$\pm$0.85 \\
SoftRankBALD & 78.80$\pm$1.93 & 86.64$\pm$1.13 & 75.52$\pm$1.11 & 77.73$\pm$1.40 \\
\hline
Top-$B$ EPIG & 84.33$\pm$1.29 & 91.11$\pm$1.02 & 78.61$\pm$0.88 & \textbf{81.27$\pm$0.80} \\
PowerEPIG & 81.53$\pm$1.39 & 87.74$\pm$0.78 & 77.24$\pm$0.64 & 79.43$\pm$0.89 \\
SoftmaxEPIG & 82.73$\pm$1.39 & 89.57$\pm$1.23 & 76.68$\pm$1.04 & 80.03$\pm$0.83 \\
SoftRankEPIG & 83.31$\pm$1.31 & 91.45$\pm$1.13 & \textbf{78.64$\pm$1.36} & 80.08$\pm$0.93 \\
\hline
ParBaLS-MAP EPIG & \textbf{85.30$\pm$0.29} & 92.23$\pm$0.87 & \underline{\textbf{79.93$\pm$1.07}} & \textbf{81.32$\pm$1.31} \\
ParBaLS EPIG & \underline{\textbf{85.78$\pm$0.70}} & \underline{\textbf{93.78$\pm$0.75}} & \textbf{79.63$\pm$1.34} & \underline{\textbf{82.13$\pm$0.62}} \\
\hline
\end{tabular}
}
\caption{Final test accuracy of Bayesian AL algorithms with Bayesian Logistic Regression on One-vs-all AG News, where each of the 10 iterations has a labeling budget of 20 samples, starting with random initialization of 20 samples.}
\label{tab:agnews_ova_bs20_bayesian_ttest}
\end{table*}

\begin{table*}
\centering
\resizebox{\columnwidth}{!}{%
\begin{tabular}{l|cccc}
\hline
Datasets & \multicolumn{4}{c}{One-vs-all AG News} \\
Algorithm & 0 & 1 & 2 & 3 \\ \hline
Top-$B$ BALD & 80.19$\pm$1.80 & 88.53$\pm$1.30 & 78.79$\pm$1.17 & 78.47$\pm$1.49 \\
PowerBALD & 85.17$\pm$1.02 & 91.46$\pm$0.81 & 80.03$\pm$0.96 & 82.53$\pm$0.67 \\
SoftmaxBALD & 85.29$\pm$0.75 & 91.45$\pm$0.82 & 79.31$\pm$1.14 & 81.87$\pm$1.01 \\
SoftRankBALD & 84.47$\pm$1.82 & 92.72$\pm$1.27 & 80.89$\pm$1.18 & 82.63$\pm$0.93 \\
\hline
Top-$B$ EPIG & \textbf{87.54$\pm$0.79} & \textbf{94.45$\pm$0.68} & \textbf{81.79$\pm$0.98} & \textbf{84.14$\pm$0.71} \\
PowerEPIG & 84.51$\pm$1.20 & 90.63$\pm$0.76 & 79.74$\pm$1.11 & 81.74$\pm$0.93 \\
SoftmaxEPIG & 84.81$\pm$1.29 & 92.29$\pm$0.33 & 80.67$\pm$1.42 & 82.89$\pm$0.71 \\
SoftRankEPIG & \textbf{87.57$\pm$0.81} & 93.92$\pm$0.57 & 81.52$\pm$0.64 & 83.32$\pm$0.90 \\
\hline
ParBaLS-MAP EPIG & \textbf{87.47$\pm$1.10} & \underline{\textbf{94.91$\pm$0.48}} & \textbf{82.67$\pm$0.53} & \textbf{84.64$\pm$0.76} \\
ParBaLS EPIG & \underline{\textbf{88.00$\pm$0.79}} & \textbf{94.83$\pm$0.39} & \underline{\textbf{83.10$\pm$1.26}} & \underline{\textbf{84.73$\pm$0.78}} \\
\hline
\end{tabular}
}
\caption{Final test accuracy of Bayesian AL algorithms with Bayesian Logistic Regression on One-vs-all AG News, where each of the 10 iterations has a labeling budget of 20 samples, starting with random initialization of 100 samples.}
\label{tab:agnews_ova_bs100_bayesian_ttest}
\end{table*}

\begin{table*}
\centering
\resizebox{\columnwidth}{!}{%
\begin{tabular}{l|cccc}
\hline
Datasets & \multicolumn{4}{c}{One-vs-all AG News} \\
Algorithm & 0 & 1 & 2 & 3 \\ \hline
Random & 80.69$\pm$0.89 & 87.59$\pm$1.17 & 76.29$\pm$1.03 & 78.02$\pm$1.32 \\
Confidence & \textbf{85.34$\pm$0.84} & 92.55$\pm$0.69 & \underline{\textbf{80.02$\pm$1.19}} & \underline{\textbf{82.35$\pm$1.22}} \\
GLISTER & 72.12$\pm$2.32 & 77.76$\pm$3.07 & 71.09$\pm$1.86 & 70.69$\pm$2.01 \\
CoreSet & 63.18$\pm$1.11 & 65.31$\pm$0.79 & 61.13$\pm$1.11 & 63.44$\pm$0.78 \\
BADGE & 81.27$\pm$0.85 & 86.75$\pm$0.86 & 75.79$\pm$1.54 & 78.40$\pm$0.85 \\
GALAXY & 84.00$\pm$1.00 & 91.94$\pm$0.66 & 78.44$\pm$0.84 & \textbf{81.39$\pm$1.16} \\
\hline
ParBaLS-MAP EPIG & \textbf{85.30$\pm$0.29} & 92.23$\pm$0.87 & \textbf{79.93$\pm$1.07} & \textbf{81.32$\pm$1.31} \\
ParBaLS EPIG & \underline{\textbf{85.78$\pm$0.70}} & \underline{\textbf{93.78$\pm$0.75}} & \textbf{79.63$\pm$1.34} & \textbf{82.13$\pm$0.62} \\
\hline
\end{tabular}
}
\caption{Final test accuracy of ParBaLS and heuristic baselines with Bayesian Logistic Regression on One-vs-all AG News, where each of the 10 iterations has a labeling budget of 20 samples, starting with random initialization of 20 samples.}
\label{tab:agnews_ova_bs20_heuristic_ttest}
\end{table*}

\begin{table*}
\centering
\resizebox{\columnwidth}{!}{%
\begin{tabular}{l|cccc}
\hline
Datasets & \multicolumn{4}{c}{One-vs-all AG News} \\
Algorithm & 0 & 1 & 2 & 3 \\ \hline
Random & 83.92$\pm$0.75 & 90.92$\pm$0.77 & 79.14$\pm$1.04 & 80.90$\pm$1.58 \\
Confidence & \textbf{87.87$\pm$0.78} & \textbf{94.82$\pm$0.76} & \textbf{82.49$\pm$1.35} & \underline{\textbf{85.08$\pm$0.78}} \\
GLISTER & 82.12$\pm$0.89 & 90.22$\pm$1.06 & 79.19$\pm$1.32 & 80.24$\pm$1.14 \\
CoreSet & 74.59$\pm$1.46 & 79.80$\pm$1.14 & 71.15$\pm$1.67 & 73.64$\pm$1.42 \\
BADGE & 83.84$\pm$0.89 & 90.65$\pm$1.06 & 79.81$\pm$0.94 & 81.15$\pm$1.31 \\
GALAXY & \textbf{87.24$\pm$0.65} & \textbf{94.23$\pm$0.64} & \textbf{81.69$\pm$0.94} & \textbf{84.17$\pm$0.79} \\
\hline
ParBaLS-MAP EPIG & \textbf{87.47$\pm$1.10} & \underline{\textbf{94.91$\pm$0.48}} & \textbf{82.67$\pm$0.53} & \textbf{84.64$\pm$0.76} \\
ParBaLS EPIG & \underline{\textbf{88.00$\pm$0.79}} & \textbf{94.83$\pm$0.39} & \underline{\textbf{83.10$\pm$1.26}} & \textbf{84.73$\pm$0.78} \\
\hline
\end{tabular}
}
\caption{Final test accuracy of ParBaLS and heuristic baselines with Bayesian Logistic Regression on One-vs-all AG News, where each of the 10 iterations has a labeling budget of 20 samples, starting with random initialization of 100 samples.}
\label{tab:agnews_ova_bs100_heuristic_ttest}
\end{table*}

\begin{table*}
\centering
\resizebox{\columnwidth}{!}{%
\begin{tabular}{l|ccccc}
\hline
Datasets & \multicolumn{5}{c}{One-vs-all Yelp} \\
Algorithm & 0 & 1 & 2 & 3 & 4 \\ \hline
Top-$B$ BALD & 63.60$\pm$1.22 & 58.83$\pm$1.17 & 60.13$\pm$1.04 & 60.87$\pm$1.04 & 65.27$\pm$0.79 \\
PowerBALD & \textbf{77.86$\pm$0.89} & 68.33$\pm$1.43 & 67.15$\pm$1.46 & 68.47$\pm$2.03 & 74.82$\pm$1.17 \\
SoftmaxBALD & 76.98$\pm$1.26 & 67.85$\pm$1.42 & 67.68$\pm$1.01 & 68.42$\pm$1.10 & 74.47$\pm$0.82 \\
SoftRankBALD & 69.87$\pm$1.31 & 64.46$\pm$1.81 & 64.86$\pm$1.32 & 65.79$\pm$0.68 & 73.24$\pm$1.63 \\
\hline
Top-$B$ EPIG & 77.31$\pm$0.84 & 68.76$\pm$1.02 & 69.06$\pm$0.83 & 68.82$\pm$0.96 & \textbf{77.09$\pm$0.93} \\
PowerEPIG & 75.76$\pm$1.01 & 68.64$\pm$1.19 & 67.16$\pm$1.34 & 66.71$\pm$1.18 & 74.19$\pm$0.88 \\
SoftmaxEPIG & 76.96$\pm$0.95 & \textbf{69.45$\pm$1.02} & \textbf{68.90$\pm$1.65} & 68.66$\pm$0.76 & 75.55$\pm$0.88 \\
SoftRankEPIG & \textbf{78.54$\pm$0.92} & \textbf{69.68$\pm$0.95} & \textbf{69.75$\pm$0.98} & \textbf{69.27$\pm$1.43} & \textbf{75.97$\pm$1.07} \\
\hline
ParBaLS-MAP EPIG & \textbf{78.41$\pm$0.45} & \textbf{69.71$\pm$1.30} & \textbf{70.46$\pm$1.54} & \textbf{70.31$\pm$0.92} & \textbf{77.04$\pm$1.00} \\
ParBaLS EPIG & \underline{\textbf{78.91$\pm$0.78}} & \underline{\textbf{70.42$\pm$0.59}} & \underline{\textbf{70.75$\pm$1.50}} & \underline{\textbf{71.01$\pm$1.48}} & \underline{\textbf{77.20$\pm$0.83}} \\
\hline
\end{tabular}
}
\caption{Final test accuracy of Bayesian AL algorithms with Bayesian Logistic Regression on One-vs-all Yelp, where each of the 10 iterations has a labeling budget of 20 samples, starting with random initialization of 20 samples.}
\label{tab:yelp_ova_bs20_bayesian_ttest}
\end{table*}

\begin{table*}
\centering
\resizebox{\columnwidth}{!}{%
\begin{tabular}{l|ccccc}
\hline
Datasets & \multicolumn{5}{c}{One-vs-all Yelp} \\
Algorithm & 0 & 1 & 2 & 3 & 4 \\ \hline
Top-$B$ BALD & 71.58$\pm$1.41 & 64.94$\pm$1.34 & 65.26$\pm$1.33 & 65.78$\pm$1.07 & 71.63$\pm$0.93 \\
PowerBALD & 78.51$\pm$1.14 & 68.51$\pm$1.27 & 69.00$\pm$0.83 & 68.79$\pm$1.24 & 76.16$\pm$1.10 \\
SoftmaxBALD & 78.11$\pm$1.22 & 70.08$\pm$1.19 & 69.26$\pm$0.95 & 69.58$\pm$1.31 & 76.28$\pm$0.68 \\
SoftRankBALD & 75.39$\pm$1.28 & 68.48$\pm$1.32 & 68.68$\pm$1.30 & 69.57$\pm$1.22 & 76.63$\pm$1.07 \\
\hline
Top-$B$ EPIG & 79.30$\pm$0.99 & 71.30$\pm$0.86 & 71.07$\pm$0.79 & \textbf{71.09$\pm$0.63} & \textbf{78.56$\pm$0.90} \\
PowerEPIG & 77.50$\pm$0.76 & 69.31$\pm$0.81 & 68.49$\pm$0.79 & 69.20$\pm$0.93 & 76.29$\pm$0.68 \\
SoftmaxEPIG & 78.56$\pm$0.90 & 70.58$\pm$1.32 & 69.62$\pm$1.28 & 70.00$\pm$0.98 & 76.86$\pm$0.66 \\
SoftRankEPIG & 79.58$\pm$0.79 & 70.69$\pm$0.95 & 70.57$\pm$1.01 & \textbf{70.60$\pm$1.20} & \textbf{77.84$\pm$0.92} \\
\hline
ParBaLS-MAP EPIG & \textbf{80.41$\pm$0.95} & \textbf{71.89$\pm$0.89} & 70.92$\pm$1.35 & \textbf{71.02$\pm$1.29} & \textbf{78.69$\pm$0.90} \\
ParBaLS EPIG & \underline{\textbf{80.89$\pm$0.54}} & \underline{\textbf{72.73$\pm$1.01}} & \underline{\textbf{72.65$\pm$1.19}} & \underline{\textbf{71.32$\pm$0.97}} & \underline{\textbf{78.76$\pm$0.77}} \\
\hline
\end{tabular}
}
\caption{Final test accuracy of Bayesian AL algorithms with Bayesian Logistic Regression on One-vs-all Yelp, where each of the 10 iterations has a labeling budget of 20 samples, starting with random initialization of 100 samples.}
\label{tab:yelp_ova_bs100_bayesian_ttest}
\end{table*}

\begin{table*}
\centering
\resizebox{\columnwidth}{!}{%
\begin{tabular}{l|ccccc}
\hline
Datasets & \multicolumn{5}{c}{One-vs-all Yelp} \\
Algorithm & 0 & 1 & 2 & 3 & 4 \\ \hline
Random & 76.47$\pm$0.83 & 68.73$\pm$1.32 & 67.67$\pm$1.42 & 66.99$\pm$1.15 & 74.00$\pm$1.12 \\
Confidence & \textbf{78.69$\pm$1.09} & 69.85$\pm$1.55 & 70.32$\pm$0.87 & \textbf{70.35$\pm$0.76} & \textbf{76.80$\pm$1.18} \\
GLISTER & 65.88$\pm$1.50 & 60.10$\pm$2.48 & 61.08$\pm$2.40 & 62.66$\pm$1.59 & 66.71$\pm$1.91 \\
CoreSet & 58.95$\pm$1.53 & 55.78$\pm$0.62 & 55.85$\pm$1.23 & 56.79$\pm$0.86 & 58.90$\pm$1.12 \\
BADGE & 75.50$\pm$0.82 & 68.10$\pm$0.96 & 66.29$\pm$0.87 & 67.83$\pm$1.17 & 74.18$\pm$0.90 \\
GALAXY & \textbf{78.51$\pm$1.05} & \underline{\textbf{71.96$\pm$1.55}} & \underline{\textbf{72.43$\pm$1.83}} & \underline{\textbf{71.16$\pm$1.40}} & \textbf{76.95$\pm$0.74} \\
\hline
ParBaLS-MAP EPIG & \textbf{78.41$\pm$0.45} & 69.71$\pm$1.30 & \textbf{70.46$\pm$1.54} & \textbf{70.31$\pm$0.92} & \textbf{77.04$\pm$1.00} \\
ParBaLS EPIG & \underline{\textbf{78.91$\pm$0.78}} & \textbf{70.42$\pm$0.59} & \textbf{70.75$\pm$1.50} & \textbf{71.01$\pm$1.48} & \underline{\textbf{77.20$\pm$0.83}} \\
\hline
\end{tabular}
}
\caption{Final test accuracy of ParBaLS and heuristic baselines with Bayesian Logistic Regression on One-vs-all Yelp, where each of the 10 iterations has a labeling budget of 20 samples, starting with random initialization of 20 samples.}
\label{tab:yelp_ova_bs20_heuristic_ttest}
\end{table*}

\begin{table*}
\centering
\resizebox{\columnwidth}{!}{%
\begin{tabular}{l|ccccc}
\hline
Datasets & \multicolumn{5}{c}{One-vs-all Yelp} \\
Algorithm & 0 & 1 & 2 & 3 & 4 \\ \hline
Random & 77.61$\pm$0.96 & 69.03$\pm$1.00 & 68.31$\pm$0.95 & 69.45$\pm$1.31 & 76.51$\pm$0.85 \\
Confidence & \textbf{80.11$\pm$1.01} & \textbf{71.63$\pm$1.03} & 71.04$\pm$0.84 & \textbf{71.44$\pm$1.00} & \textbf{78.72$\pm$0.97} \\
GLISTER & 73.07$\pm$1.02 & 68.39$\pm$1.41 & 66.46$\pm$1.92 & 68.57$\pm$1.01 & 74.49$\pm$1.26 \\
CoreSet & 69.26$\pm$1.12 & 63.23$\pm$1.27 & 64.07$\pm$1.29 & 64.56$\pm$1.17 & 69.09$\pm$1.23 \\
BADGE & 77.94$\pm$1.23 & 68.74$\pm$0.80 & 68.96$\pm$1.39 & 68.40$\pm$0.92 & 75.88$\pm$1.29 \\
GALAXY & \textbf{80.28$\pm$0.79} & \underline{\textbf{72.74$\pm$0.89}} & \underline{\textbf{72.73$\pm$1.08}} & \underline{\textbf{72.66$\pm$1.19}} & \underline{\textbf{78.91$\pm$0.87}} \\
\hline
ParBaLS-MAP EPIG & \textbf{80.41$\pm$0.95} & \textbf{71.89$\pm$0.89} & 70.92$\pm$1.35 & 71.02$\pm$1.29 & \textbf{78.69$\pm$0.90} \\
ParBaLS EPIG & \underline{\textbf{80.89$\pm$0.54}} & \textbf{72.73$\pm$1.01} & \textbf{72.65$\pm$1.19} & \textbf{71.32$\pm$0.97} & \textbf{78.76$\pm$0.77} \\
\hline
\end{tabular}
}
\caption{Final test accuracy of ParBaLS and heuristic baselines with Bayesian Logistic Regression on One-vs-all Yelp, where each of the 10 iterations has a labeling budget of 20 samples, starting with random initialization of 100 samples.}
\label{tab:yelp_ova_bs100_heuristic_ttest}
\end{table*}

\section{Licenses of Software and Datasets}\label{sec:licenses}

\subsection{Software}
 
Our code builds on LabelBench \citep{zhang2024labelbench} (MIT License), which provides the Random, Confidence, BADGE, CoreSet, and GALAXY implementations; all other acquisition strategies are our own implementations. Bayesian logistic regression uses PyMC \citep{AbrilPla2023PyMCAM} (Apache License 2.0). Embeddings are computed with CLIP ViT-B/32 \citep{radford2021learning} (MIT License), DINOv2 ViT-S/14 \citep{oquab2024dinov2} (Apache License 2.0), and bert-base-uncased \citep{devlin2019bert} (Apache License 2.0).
 
\subsection{Datasets}
 
\begin{itemize}
    \item Airline Passenger Satisfaction \citep{airline2020}: no license specified.
    \item Credit Card Fraud \citep{credit2022}: CC0 1.0.
    \item Magic Gamma Telescope, Statlog Landsat Satellite, and Vowels \citep{Kelly2025UCI}: CC BY 4.0.
    \item Civil Comments \citep{DBLP:journals/corr/abs-1903-04561}: CC0 1.0.
    \item AG News \citep{agnews2004}: research and non-commercial use; no formal license.
    \item Yelp Review Full \citep{zhang2015character}: Yelp Dataset Agreement (academic, non-commercial use).
    \item CIFAR-10 \citep{krizhevsky2009learning}: no license specified.
    \item iWildCam \citep{beery2021iwildcam, koh2021wilds}: CDLA-Permissive-1.0.
    \item fMoW \citep{christie2018functional, koh2021wilds}: FMoW Challenge Public License (non-commercial).
\end{itemize}

\end{document}